\documentclass[10pt,journal,compsoc]{IEEEtran}

\usepackage[utf8]{inputenc} 
\usepackage[T1]{fontenc}   
\usepackage[colorlinks]{hyperref}       
\usepackage{url}            
\usepackage{booktabs}       
\usepackage{amsfonts}      
\usepackage{nicefrac}    
\usepackage{microtype}     

\newcommand{\ie}{\textit{i}.\textit{e}.}
\usepackage{amsmath}
\usepackage{multirow}
\usepackage[table,xcdraw]{xcolor}
\usepackage{bm}
\usepackage{paralist} 
\usepackage{graphicx}
\usepackage{wrapfig}
\usepackage{bbding}
\usepackage{multicol}
\usepackage{color}
\usepackage{amssymb}
\usepackage{cite}
\usepackage[sort,numbers]{natbib}
\usepackage{amsthm,amsmath,amssymb}
\usepackage{mathrsfs}
\usepackage{stfloats}
\usepackage{amssymb}
\usepackage{adjustbox}

\usepackage[percent]{overpic}

\makeatletter
\newcommand\notsotiny{\@setfontsize\notsotiny{7}{7.1828}}
\makeatother
\newcommand{\Paragraph}[1]{\noindent\textbf{#1}}

\def\etal{\emph{et al}.~}
\def\ie{i.e.} 
\def\eg{\emph{e.g.}}

\def\blu#1{\textbf{\color{blue} #1}} 
\def\red#1{\textbf{\color{red}\underline{#1}}} 
\def\cite{\citep}

\begin{document}

\title{VST\texttt{++}: Efficient and Stronger Visual Saliency Transformer}

\author{Nian~Liu,
        Ziyang~Luo,
        Ni~Zhang,
        and~Junwei~Han,~\IEEEmembership{Fellow,~IEEE}

\IEEEcompsocitemizethanks{\IEEEcompsocthanksitem N. Liu is with the Computer Vision Department, Mohamed bin Zayed University of Artificial Intelligence, Abu Dhabi, UAE. (e-mail: liunian228@gmail.com)
\IEEEcompsocthanksitem Z. Luo, N. Zhang, and J. Han are with School of Automation, Northwestern Polytechnical University, Xi'an, China, 710072, E-mail: \{ziyangluo1110, nnizhang.1995 and junweihan2010\}@gmail.com
\IEEEcompsocthanksitem Junwei Han is the corresponding author.}}

\markboth{Journal of \LaTeX\ Class Files,~Vol.~14, No.~8, August~2015}%
{Shell \MakeLowercase{\textit{et al.}}: Learning Non-target Knowledge for Few-shot semantic segmentation}

\IEEEtitleabstractindextext{
\begin{abstract}

    While previous CNN-based models have exhibited promising results for salient object detection (SOD), their ability to explore global long-range dependencies is restricted.
    Our previous work, the Visual Saliency Transformer (VST), addressed this constraint from a transformer-based sequence-to-sequence perspective, to unify RGB and RGB-D SOD.
    In VST, we developed a multi-task transformer decoder that concurrently predicts saliency and boundary outcomes in a pure transformer architecture.
    Moreover, we introduced a novel token upsampling method called reverse T2T for predicting a high-resolution saliency map effortlessly within transformer-based structures.
    Building upon the VST model, we further propose an efficient and stronger VST version in this work, \ie VST\texttt{++}. 
    To mitigate the computational costs of the VST model, we propose a Select-Integrate Attention (SIA) module,  partitioning foreground into fine-grained segments and aggregating background information into a single coarse-grained token. 
    To incorporate 3D depth information with low cost, we design a novel depth position encoding method tailored for depth maps. 
    Furthermore, we introduce a token-supervised prediction loss to provide straightforward guidance for the task-related tokens.
    We evaluate our VST\texttt{++} model across various transformer-based backbones on RGB, RGB-D, and RGB-T SOD benchmark datasets. Experimental results show that our model outperforms existing methods while achieving a 25\% reduction in computational costs without significant performance compromise. The demonstrated strong ability for generalization, enhanced performance, and heightened efficiency of our VST\texttt{++} model highlight its potential.
\end{abstract}

}

\maketitle
\IEEEdisplaynontitleabstractindextext

\IEEEpeerreviewmaketitle

\IEEEraisesectionheading{\section{Introduction}\label{sec1}}
\IEEEPARstart{S}{alient} object detection (SOD) aims at identifying and segmenting the most prominent objects in images.
It can be utilized as a pre-processing technique to promote various computer vision tasks, such as weakly-supervised learning \cite{shimoda2016distinct, zhang2022generalized}, person re-identification \cite{zhao2016person}, and video object segmentation \cite{wang2015saliency,lu2020zero,wang2020paying}.
Apart from RGB images, depth data captured by depth cameras provides valuable spatial structure information. RGB-D SOD capitalizes on these two data types and has garnered increasing attention in recent years.

Previous cutting-edge RGB and RGB-D SOD methods were mainly dominated by convolutional networks \cite{lecun1998gradient}, \ie\ using encoder-decoder CNN architectures \cite{noh2015learning,ronneberger2015unet}, where the encoder extracts multi-level features from the input image via pre-trained backbones \cite{simonyan2014vgg, he2016resnet} and the decoder fuses the encoded features for saliency prediction.
Based on this effective architecture, previous works mainly focused on designing powerful decoders to accurately generate saliency maps. To this end, they have proposed various multi-scale feature aggregation approaches \cite{hou2018dss,MINet-CVPR2020,fan2020bbsnet,luo2020Cas-Gnn}, multi-task learning schemes \cite{wang2018salient,zhang2019capsal,zhao2019EGNet,CVPR2020_LDF,Wei2020CoNet}, and attention mechanisms \cite{liu2018picanet,zhang2018pagr,chen2020dpanet}.
For RGB-D SOD, a supplementary challenge involves effectively fusing cross-modal cues, \ie\ the RGB appearance information and the depth information. Existing works have presented lots of cross-modal fusion approaches, such as attention models \cite{Li2020CMWNet,zhang2020ATSA}, feature integration \cite{han2017cnns,chen2018progressively,fan2020bbsnet,Fu2020JLDCF,specificity_rgbd_sod}, dynamic convolution \cite{HDFNet-ECCV2020}, graph neural networks \cite{luo2020Cas-Gnn}, and knowledge distillation \cite{piao2020a2dele}.
Consequently, CNN-based approaches have achieved promising results, as summarized in \cite{wang2019salient1,zhou2021rgb}.

Nevertheless, due to intrinsic constraints of CNNs, previous CNN-based models have limited modeling capabilities for exploring global long-range dependencies, which has proven as a crucial factor for saliency detection, such as global contexts \cite{goferman2011context,zhao2015saliency,ren2015exploiting,luo2017non,liu2018picanet} and global contrast \cite{zhai2006visual,borji2012exploiting,cheng2014global}.
Although fully connected layers \cite{liu2016dhsnet,han2017cnns}, global pooling layers \cite{luo2017non,liu2018picanet,wang2017stagewise}, and non-local modules \cite{liu2020S2MA,chen2020dpanet} have been proposed to alleviate this issue, they are merely inserted as certain modules to explore global cues, while the standard CNN-based architecture remains unchanged.

For tackling the limitation of CNNs, \cite{vaswani2017attention} presented the Transformer architecture to model global long-range dependencies among word sequences for machine translation. The key element, \ie\ self-attention, computes the query-key correlation across the sequence and is cascaded layer by layer in both encoder and decoder, thereby allowing for the learning of global long-range dependencies between different positions in every layer. 
With this in mind, it is natural to introduce the Transformer to SOD for employing global cues thoroughly.

In our ICCV 2021 version of this work \cite{Liu_2021_ICCV}, we rethought SOD from a new sequence-to-sequence point of view and proposed the first pure transformer model based on the recently proposed Vision Transformer (ViT) models \cite{dosovitskiy2020image,yuan2021tokens} to tackle both RGB and RGB-D SOD tasks, namely Visual Saliency Transformer (VST).
In ViT models \cite{dosovitskiy2020image,yuan2021tokens}, each image is initially divided into patches as a patch sequence, which is further fed into the Transformer model to handle global long-range dependencies across the whole image, without relying on any convolution operations.
However, we would encounter two problems when following this perspective to introduce ViT for SOD.
On the one hand, SOD is a dense prediction task that differs from image classification solved by the ViT model, thus raising an open question of how to perform dense prediction tasks based on pure transformer. On the other hand, ViT tokenizes the image at a coarse scale which contrasts with the fine-grained requirements of SOD. Hence, it is also unclear how to adapt ViT to the high-resolution prediction demand of SOD.

To address the first problem, we presented a token-based transformer decoder by designing task-related tokens to acquire task-related decision embeddings. Subsequently, these tokens were employed to generate dense predictions by the proposed novel patch-task-attention mechanism, providing a fresh approach to applying transformers in dense prediction tasks. Furthermore, drawing inspiration from previous SOD models \cite{zhao2019EGNet,Zhou2020ITSD,zhang2020select,Wei2020CoNet} that utilized boundary detection to improve SOD performance, we constructed a multi-task decoder by incorporating saliency and boundary tokens to concurrently handle saliency and boundary prediction tasks. This approach streamlined the multitask prediction process by focusing on learning task-specific tokens, resulting in significant reductions in computational overhead while achieving improved results.
To solve the second problem, motivated by the Tokens-to-Token (T2T) transformation \cite{yuan2021tokens} that decreases the lengths of tokens, we introduced a novel reverse T2T transformation for token upsampling. This innovative approach progressively upsampled patch tokens by expanding each token into multiple sub-tokens. Subsequently, the upsampled patch tokens were fused with low-level tokens to generate a complete full-resolution saliency map.
Besides, a cross-modality transformer was employed to thoroughly excavate the interaction between RGB and depth cues for RGB-D SOD. Ultimately, our VST proved to hold superior performance compared with state-of-the-art SOD methods on both RGB and RGB-D SOD benchmark datasets.

Despite the solutions to the aforementioned problems that facilitate the application of ViT to the SOD task, the VST model still exhibits several limitations. To this end, we propose a new improved version, \ie\ VST\texttt{++}, to overcome these shortcomings from the perspectives of efficiency, performance, and generalization.
First, the quadratic computation required for self-attention poses a significantly high computational cost for VST.
To improve model efficiency, we propose a Select-Integrate Attention (SIA) module, building upon recent developments in coarse-grained global attention and fine-grained local attention approaches \cite{liu2021Swin, vaswani2021scaling, zhang2021multi}. 
For the SOD task, foreground region is necessary for accurate segmentation, while the background region also provides useful contextual cues, however, it is 
redundant. To this end, we propose to select the foreground regions, keeping their fine-grained segments, and integrate background regions into a coarse-grained global token.
In this way, we preserve the accuracy of foreground information and also avoid computational costs on redundant background regions.
Second, traditional sinusoidal position encoding (PE) used in Transformers solely considers the $x$ and $y$ dimensions, disregarding crucial depth information for RGB-D data. To address this limitation, we design a new depth position encoding tailored for depth maps, hence introducing 3D depth cues into the decoder in a simple and lightweight way.
Third, we improve the model performance by introducing an additional loss function. From the results of VST, we observed that 
the predictions heavily rely on patch tokens and the saliency and boundary tokens only encoded auxiliary task-related information. We hypothesize this issue stems from the lack of direct supervision on the saliency and boundary tokens. 
Hence, we propose a token-supervised prediction loss to provide straightforward supervision for the task-related tokens.
Fourth, we verify the generalization ability of VST\texttt{++} with different transformer backbones and on a new multi-modality SOD task, \ie\ RGB-T SOD.
These comprehensive experiments unequivocally showcase the capabilities of our VST\texttt{++} model across different network scales and data modalities.

In our previous VST model \cite{Liu_2021_ICCV}, we mainly had three contributions.
\begin{compactitem}
\item 
To the best of our knowledge, our VST stood as the pioneering unified model employing a pure transformer architecture for both single and multi-modality-based SOD from a new perspective of sequence-to-sequence modeling.

\item We proposed a multi-task transformer decoder to simultaneously perform saliency and boundary detection by designing task-related tokens and patch-task-attention.

\item In lieu of the commonly used bilinear upsampling in CNN architectures, we introduced a reverse T2T token upsampling method under the pure transformer-based framework and showcased its efficacy.
\end{compactitem}

In this work, we further make the following contributions:
\begin{compactitem}
\item We propose an SIA module by partitioning foreground into fine-grained segments and aggregating background information into a single coarse-grained token. This approach decreases computational costs by 25\% without significantly compromising performance.

\item We introduce a novel depth position encoding for RGB-D SOD as a supplementary to the traditional 2D spatial PE, introducing depth cues in the decoder in a simple way.

\item We introduce a token-supervised prediction loss to further enhance the model performance by providing direct supervison for task-related tokens.

\item We further verify the effectiveness of our proposed VST\texttt{++} model using different backbone architectures and on the RGB-T SOD task. Our final results show new state-of-the-art performance on several widely used RGB, RGB-D, and RGB-T SOD benchmark datasets, signifying the immense potential of transformer-based models for SOD.
\end{compactitem}

\section{Related Work}

\subsection{CNN-Based SOD}
CNN-based methods have become a prevailing trend not only in the segmentation field \cite{fang2023reliable}, delivering impressive performances owing to the powerful feature learning ability of CNNs.
RGB SOD methods can be broadly classified into four groups: attention-based, multi-level fusion-based, recurrent-model-based, and multi-task-based approaches.
Multi-level fusion-based approaches \cite{hou2018dss,wang2017stagewise,MINet-CVPR2020,GateNet,fan2020bbsnet} often utilized UNet \cite{ronneberger2015unet} and HED-style \cite{xie2015hed} networks as a foundation to aggregate both high-level and low-level features.
Besides, some works \cite{wang2017stagewise,luo2020Cas-Gnn} adopted spatial pyramid pooling and \cite{GateNet,fan2020bbsnet} applied the dilated convolution \cite{chen2017deeplab} for gathering multi-scale features.
Attention-based methods \cite{Piao2019dmra,zhang2018pagr,fan2020bbsnet,chen2020dpanet} and \cite{liu2018picanet} effectively selected discriminative features by resorting to spatial, channel, or pixel-wise contextual attention.
Recurrent-model-based works \cite{liu2016dhsnet,wang2018rfcn,deng2018r3net,liu2019salient,chen2020PGAR} focused on refining the details of saliency map step by step.
Multi-task-based methods tried to 
introduce other tasks, such as fixation prediction \cite{wang2018salient}, image caption \cite{zhang2019capsal}, and edge detection \cite{qin2019basnet, zhao2019EGNet,CVPR2020_LDF,zhang2020select,Wei2020CoNet}, to enhance SOD performance. For a more comprehensive literature review, please refer to \cite{wang2019salient1}.
For RGB-D SOD, an effective fusion of RGB and depth features is critical.
For simplicity, early works \cite{chen2018progressively, chen2019three, Fu2020JLDCF} utilized basic operations (\eg, concatenation, summation, or multiplication) to fuse information from two modalities.
Recent studies \cite{zhao2019contrast, li2020icnet, Piao2019dmra, Li2020CMWNet} leveraged attention mechanism such as spatial or channel attention generated from depth cues to improve the RGB features.
Additionally, dynamic convolution \cite{HDFNet-ECCV2020}, graph neural networks \cite{luo2020Cas-Gnn}, and knowledge distillation \cite{piao2020a2dele} were also used to perform multi-modal feature fusion.
As a counterpart, cross-attention was invoked in \cite{liu2020S2MA, liu2021learning, chen2020dpanet} to propagate long-range cross-modality cues. Please refer to \cite{zhou2021rgb} for a more comprehensive literature review.

Another multi-modal SOD task, \ie\ RGB-T SOD, also aims to explore complementary cues for accurate salient object detection. 
Various traditional graph-based approaches have been proposed to obtain low-level hand-crafted features \cite{wang2018rgb, tu2019m3s, tu2019rgb}. 
Recently, 
some works have used CNNs to extract image features and proposed diverse fusion methods, including multi-level feature fusion \cite{zhang2019rgb, tu2021multi}, spatial-wise attention mechanism \cite{tu2022rgbt}, and channel attention mechanism module \cite{wang2021cgfnet}.
Moreover, CNN-based architectures have been introduced in the context of light field SOD \cite{fu2022light, chen2023fusion}, deep unsupervised SOD \cite{liu2024deep}, and Part-Object Relational Saliency \cite{liu2022disentangled}.

Unlike earlier CNN-based methods, we introduce a fresh viewpoint from a sequence-to-sequence perspective and propose a unified model based on pure transformer architecture
for both RGB and multi-modal SOD. 
Inspired by the excellent achievements of multi-task learning, we introduce boundary detection \cite{qin2019basnet, zhao2019EGNet,CVPR2020_LDF,zhang2020select,Wei2020CoNet} through our proposed novel token-based multi-task decoder, which is noticeably distinct from previous CNN-based methods.

\subsection{Transformers in Computer Vision}
Vaswani \etal \cite{vaswani2017attention} initially proposed a transformer encoder-decoder architecture for machine translation. Since then, the transformer model has been successfully applied to several computer vision tasks, yielding impressive results. 
This success has led to the emergence of hybrid architectures that integrate CNNs and transformers in tasks such as object detection \cite{carion2020end,zhu2020deformable}, panoptic segmentation \cite{wang2020maxdeeplab}, lane shape prediction \cite{liu2021end} and so on. These hybrid models commonly adopt CNNs for feature extraction and use transformers to capture long-range dependencies.

Additionally, pure transformer models were proposed in \cite{dosovitskiy2020image} for image classification from the sequence-to-sequence perspective, which involved dividing each image into a sequence of flattened 2D patches and then applying transformers.
Subsequently, 
Wang \etal \cite{wang2021pvt} presented a pyramid architecture to apply ViT for dense prediction tasks.
T2T-ViT \cite{yuan2021tokens} employed the T2T module to generate multiscale token features by modeling local structures.
Building on this work, we design a reverse T2T token upsampling method under the pure transformer setting. 
In traditional vision transformers \cite{dosovitskiy2020image,touvron2020training,deng2009imagenet}, class token is introduced for image classification. 

\begin{figure*}[!t]
  \graphicspath{{Figures/Network/}}
  \centering
  \includegraphics[width=1\linewidth]{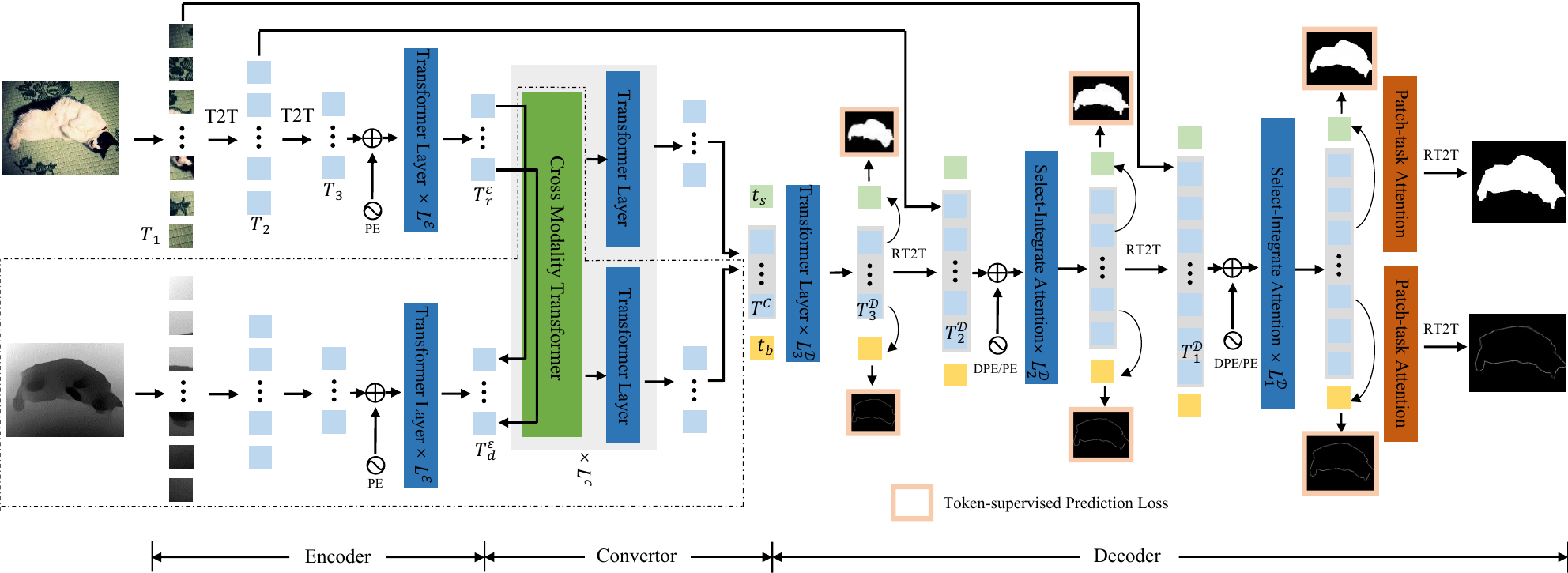}
  \caption{
\textbf{Overall architecture of our proposed VST\texttt{++} model for both RGB and RGB-D SOD.} The encoder generates multi-level tokens from the input image patch sequence. Next, the convertor projects the patch tokens to the decoder space and performs cross-modal information fusion for RGB-D SOD. 
Finally, a multi-task transformer decoder simultaneously performs saliency detection and
boundary segmentation via the proposed task-related tokens and the patch-task-attention mechanism. 
An RT2T transformation is designed to progressively upsample the patch tokens.
{In 1/8 and 1/4 decoder levels, we design Select-Integrate Attention (SIA) to select fine-grained foreground segments and aggregate coarse-grained background information for low-cost attention computation. We maintain the self-attention mechanism for the 1/16 decoder level, given the absence of a mask from the previous stage.} Depth Position Embedding (DPE) or sinusoidal position encoding (PE) is added to the query and key of the SIA for RGB-D SOD and RGB SOD, respectively. Additionally, we calculate the token-supervised prediction losses at every decoder level.
The dotted line represents components exclusively designed for RGB-D SOD.
 }
  \label{VST}
\end{figure*}
However, this can not be directly employed for generating dense predictions for SOD. Thus, we introduce a saliency and a boundary token and propose to conduct patch-task-attention for predicting saliency and boundary maps. Concurrently, we adopt a token-supervised prediction loss to supervise the task information embedding ability of these task tokens.

In the Transformer model, high-resolution vision tasks can pose significant computational challenges due to the quadratic computation involved. To address this, recent works have explored alternative approaches such as employing coarse-grained global self-attention or fine-grained local self-attention to mitigate the computational burden. For instance, Liu \etal \cite{liu2021Swin} introduced window attention for local regions and window shifts for global regions, resulting in a significant reduction in computational cost. Vaswani \etal \cite{vaswani2021scaling} proposed blocked local attention and attention downsampling methods that resemble convolution. Zhang \etal \cite{zhang2021multi} developed a 2-D version of Longformer \cite{beltagy2020longformer}, which utilized both sparsity and memory mechanisms. Yang \etal \cite{yang2021focal} designed a variant attention method that takes into account the distance between each token and its surrounding tokens. In this paper, we introduce a concept of combining coarse-grained and fine-grained attention in our decoder, namely SIA, by partitioning foreground regions into fine-grained segments while aggregating background information into one global token to reduce computational expenses and maintain accuracy. Through these techniques, we achieve a 25\% reduction in computational costs without compromising much precision in our approach.

\subsection{Transformer-Based SOD}
Recently, numerous transformer-based SOD techniques have emerged, demonstrating remarkable achievements. In the realm of RGB SOD, ICON \cite{zhuge2021salient} effectively learned integrity features from both micro and macro levels by aggregating diverse features, enhancing feature channels, and identifying the agreement of part and whole object features. EBMGSOD \cite{jing_ebm_sod21} introduced an energy-based prior for SOD and employed the Markov chain Monte Carlo-based maximum likelihood estimation to concurrently train vision transformer network and energy-based prior model. For RGB-D SOD, SwinNet \cite{liu2021swinnet} successfully unified RGB-D SOD and RGB-T SOD, aligning spatial and re-calibrating channels with the help of hierarchical features generated by the Swin Transformer \cite{liu2021Swin}. HRTransNet \cite{tang2022hrtransnet} also unified these two tasks and applied a high-resolution network \cite{yuan2021hrformer} to SOD.

In contrast to these transformer-based SOD models, we tackle the challenge from the perspective of increased efficiency and propose a SIA module, which is specifically designed for the SOD task to solve the computational costs of self-attention. Furthermore, our VST\texttt{++} is a pure transformer-based method, setting it apart from the aforementioned techniques.

\subsection{{Generic Dense Segmentation Architecture}}
{Recently, several generic dense prediction frameworks have emerged for a range of segmentation tasks. 
For instance, K-Net \cite{zhang2021k} introduced distinct dynamic kernels tailored for class-specific segmentation. Within DETR-based models, SOLQ \cite{dong2021solq} employed object-specific queries to predict class, location, and mask. CMT-DeepLab \cite{yu2022cmt} treated object queries as cluster centers, facilitating the grouping of pixels for segmentation. In Panoptic SegFormer \cite{li2022panoptic}, query responsibilities were segregated, directing them to the location decoder and mask decoder. Mask2Former \cite{cheng2022masked}, building on MaskFormer \cite{cheng2021per}, integrated masked attention to achieve a universal segmentation design. Expanding upon Mask2Former, FastInst \cite{he2023fastinst} and MP-former \cite{zhang2023mp} were introduced, featuring innovatively designed instance activation-guided queries and a mask-piloted training approach.}

{While we acknowledge the superiority of generic dense prediction models, it's important to highlight a substantial gap when applying them to SOD tasks owing to the lack of class information. Unlike the approaches mentioned above, our VST++ designed modules specifically for SOD tasks.
For example, our multi-task decoder places emphasis on boundary detection, taking into account the characteristics of the SOD task. The introduction of SIA is geared towards diminishing the computational overhead associated with self-attention, by leveraging the nature of SOD. Additionally, we put forward the DPE method, specifically tailored to encode depth position for RGB-D SOD.
These designs helps us achieve superior performance on the SOD task compared with existing general models.}

\section{VST\texttt{++}}
As depicted in Figure~\ref{VST}, the proposed VST\texttt{++} model consists of three crucial components, \ie\ a transformer encoder that depends on T2T-ViT, a transformer convertor that converts patch tokens from the encoder space to the decoder space, and a multi-task transformer decoder with Select-Integrate Attention (SIA) modules to reduce computational costs.

\subsection{Transformer Encoder}
Pretrained image classification models  (\eg, VGG \cite{simonyan2014vgg} and ResNet \cite{he2016resnet}) are commonly used as feature extractors in CNN-based SOD methods. 
In this work, any pretrained transformer backbone can be used as our encoder, such as T2T-ViT \cite{yuan2021tokens} and SwinTransformer \cite{liu2021Swin}. Here we take T2T-ViT as our default backbone encoder and detail its architecture below.

\subsubsection{Tokens to Token}
Given an input image $\bm{I}\in{\mathbb{R}^{h\times{w\times c}}}$, where $h$, $w$, and $c$ represent the height, width, and channel number, respectively, the T2T-ViT originally embeds $\bm{I}$ into a sequence of patch tokens $\bm{T}_{0}$ with a length of $l_{0}$ using a soft split operation. Subsequently, the patch tokens undergo a sequence of Tokens to Token (T2T) modules, each of which consists of a re-structurization operation and a soft split operation.

\begin{figure*}[!t]
  \graphicspath{{Figures/Network/}}
  \centering
  \includegraphics[width=1\linewidth]{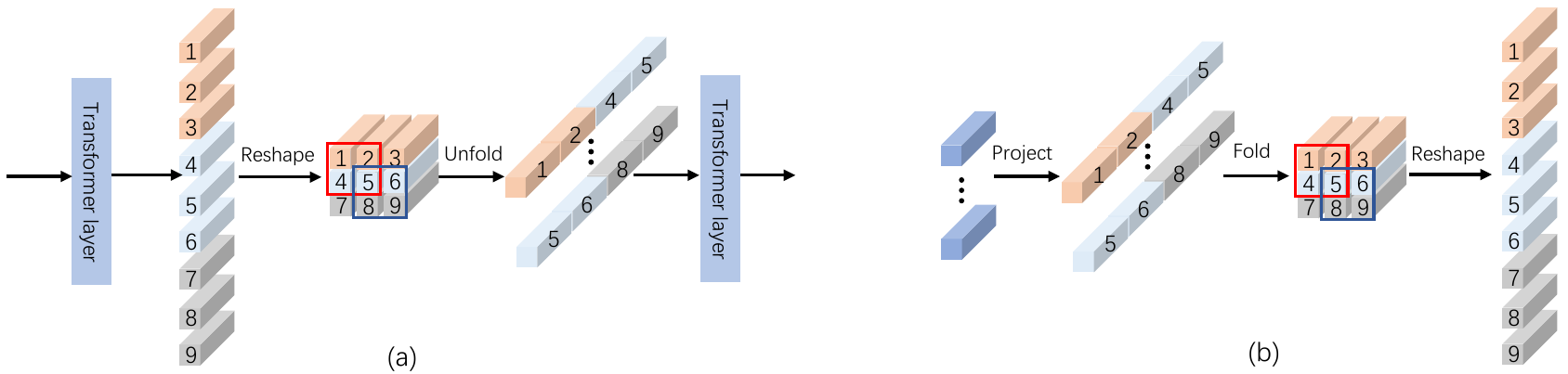}
  \caption{
  \textbf{(a) T2T module merges adjacent tokens into a new token, effectively reducing token length. (b) Our reverse T2T module enlarges each token into multiple sub-tokens, achieving token upsampling.}}
  \label{T2T}
\end{figure*}

\vspace{2mm}
\Paragraph{Re-structurization.}
During the re-structurization step, the input tokens $\bm{T}_{i} \in 
{\mathbb{R}^{l_{i}\times c}}$ 
undergo both multi-head attention and a multilayer perceptron to yield a new token sequence $\bm{T}'_{i} \in {\mathbb{R}^{l_{i}\times c}}$, where $i$ indicates the index of T2T modules. Layer normalization is applied before each block. To recover the spatial structure, $\bm{T}'_{i}$ is reshaped into a 2D image $\bm{I}_{i}\in{\mathbb{R}^{h_{i}\times{w_{i}\times c}}}$, where $l_{i} = h_{i} \times w_{i}$, as displayed in Figure~\ref{T2T}(a):
\begin{equation} \label{reconstruction}
\bm{T}'_{i} = \text{MLP}(\text{MSA}(\bm{T_{i}})).
\end{equation}
Here, MSA indicates the multi-head self-attention and MLP denotes the multilayer perceptron in the original Transformer \cite{vaswani2017attention}.

\vspace{2mm}
\Paragraph{Soft split.}
As for the soft split step, $\bm{I}_{i}$ is partitioned into $k\times k$ patches with $s$ overlapping pixels. Zero-padding of $p$ pixels is also applied to pad image boundaries. Afterward, the image patches are unfolded into a sequence of tokens $\bm{T}_{i+1} \in {\mathbb{R}^{l_{i+1}\times ck^2}}$, where the length of the sequence, $l_{i+1}$, is defined as follows:
\begin{flalign} \label{softsplit}
\begin{aligned}
\begin{split}
l_{i+1} 
&=h_{i+1}\times w_{i+1} \\
&=\!\!\lfloor \frac{h_{i}+2p-k}{k-s}\!\!+\!\!1\rfloor\!\!\times\!\!\lfloor \frac{w_{i}+2p-k}{k-s}\!\!+\!\!1\rfloor. \!\!
\end{split}
\end{aligned}
\end{flalign}
The introduction of overlapped patch splitting incorporates spatial priors by establishing correlations among neighboring patches, which is different from the original ViT \cite{dosovitskiy2020image}.

The T2T transformation can be iterated multiple times. At each iteration, the re-structurization step first converts the preceding embeddings into new embeddings, establishing long-range dependencies among all tokens. During the soft split operation, tokens are merged into a new token within each $k \times k$ neighbor and will be used in the next layer. The length of tokens can be gradually decreased by setting $s < k-1$.

Following \cite{yuan2021tokens}, we first softly split the input images into patches and then apply the T2T module twice. The overlappings are set to $s = [3, 1, 1]$, the patch sizes are defined as $k = [7, 3, 3]$, and the padding sizes are specified as $p = [2, 1, 1]$ for the three soft split steps.
After these three steps, we can acquire multi-level tokens, namely $\bm{T}_1 \in {\mathbb{R}^{l_1\times c}}$, $\bm{T}_2 \in {\mathbb{R}^{l_2\times c}}$, and $\bm{T}_3 \in {\mathbb{R}^{l_3\times c}}$, where $l_1 = h/4 \times w/4$, $l_2 = h/8 \times w/8$, and $l_3 = h/16 \times w/16$. As stated in \cite{yuan2021tokens}, we utilize a linear projection layer on $\bm{T}_3$ to alter its embedding dimension from $c=64$ to $d=384$.

\vspace{3mm}
\subsubsection{Encoder with T2T-ViT Backbone}
To encode 2D position information, we add the sinusoidal position embedding on the final token sequence $\bm{T}_3$. Furthermore, $L^{\mathcal{E}}$ transformer layers are utilized to establish long-range dependencies among $\bm{T}_3$, resulting in the creation of robust patch token embeddings $\bm{T}^{\mathcal{E}} \in {\mathbb{R}^{l_3\times d}}$.

For RGB SOD, we employ a single transformer encoder to derive RGB encoder patch tokens $\bm{T}_r^{\mathcal{E}} \in {\mathbb{R}^{l_3\times d}}$ from each input RGB image.
In the case of RGB-D SOD, we use two-stream architectures to incorporate an additional transformer encoder. This encoder is responsible for extracting the depth encoder patch tokens $\bm{T}_d^{\mathcal{E}}$ from the input depth map, following a similar procedure as the RGB encoder. An overview of this process is depicted in Figure~\ref{VST}.

\vspace{3mm}
\subsection{Transformer Convertor}
Between the transformer encoder and decoder, we introduce a convertor module that converts the encoder patch tokens $\bm{T}_*^{\mathcal{E}}$ from the encoder space to the decoder space. As a result, we obtain the converted patch tokens $\bm{T}^{\mathcal{C}} \in {\mathbb{R}^{l_3 \times d}}$, which can then be utilized in the decoder for subsequent processing.

\vspace{3mm}
\subsubsection{RGB-D Convertor}
To integrate the complementary information between the RGB and depth data, we develop a Cross Modality Transformer (CMT), which comprises $L^{\mathcal{C}}$ alternating cross-modality-attention layers and self-attention layers, to fuse $\bm{T}_r^{\mathcal{E}}$ and $\bm{T}_d^{\mathcal{E}}$ in the RGB-D converter.

\vspace{2mm}
\Paragraph{Cross-modality-attention.}
Following the pure transformer architecture, a cross-modality-attention layer is utilized to explore long-range cross-modal dependencies between the image and depth data.
The cross-modality attention follows the format of the self-attention as described in \cite{vaswani2017attention}, however, generating query, key, and value from different modalities. 
Specifically, we use the $\bm{T}_d^{\mathcal{E}}$ to generate key and value for enhancing the $\bm{T}_r^{\mathcal{E}}$, and use the $\bm{T}_r^{\mathcal{E}}$ to generate key and value for enhancing the $\bm{T}_d^{\mathcal{E}}$:

\begin{equation} \label{cross-at}
\begin{split}
\text{CA}(\bm{T}_r^{\mathcal{E}}, \bm{T}_d^{\mathcal{E}}) = \delta(\frac{\bm{T}_r^{\mathcal{E}}\bm{W}_q(\bm{T}_d^{\mathcal{E}}\bm{W}_k)^{\top}} {\sqrt{d}})\bm{T}_d^{\mathcal{E}}\bm{W}_v, \\
\text{CA}(\bm{T}_d^{\mathcal{E}}, \bm{T}_r^{\mathcal{E}}) = \delta(\frac{\bm{T}_d^{\mathcal{E}}\bm{W}_q(\bm{T}_r^{\mathcal{E}}\bm{W}_k)^{\top}} {\sqrt{d}})\bm{T}_r^{\mathcal{E}}\bm{W}_v,
\end{split}
\end{equation}
where $\delta$ means the softmax function. $\bm{W}_q,\bm{W}_k,\bm{W}_v \in {\mathbb{R}^{d\times d}}$ are projection weights for query, key, and value, respectively.

We adopt the standard transformer architecture described in \cite{vaswani2017attention} and use the multi-head attention mechanism in the cross-modality-attention. Our CMT layer also incorporates the same position-wise feed-forward network, residual connections, and layer normalization \cite{ba2016layer}.

Following the CMT, a standard transformer layer is applied to each RGB and depth patch token sequence to enhance their token embeddings. Finally, as presented in Figure~\ref{VST}, we concatenate the obtained RGB tokens and depth tokens and then project them to the final converted tokens $\bm{T}^{\mathcal{C}}$.

\vspace{3mm}
\subsubsection{RGB Convertor}
To synchronize with our RGB-D SOD model, we straightforwardly utilize $L^{\mathcal{C}}$ standard transformer layers on $\bm{T}_r^{\mathcal{E}}$ for RGB SOD, thereby generating the converted patch token sequence $\bm{T}^{\mathcal{C}}$.

\vspace{3mm}
\subsection{Multi-task Transformer Decoder}
The goal of our decoder is to decode the patch tokens $\bm{T}^{\mathcal{C}}$ to generate saliency maps. 
To accomplish this, we present a novel token upsampling method with multi-level token fusion and a token-based multi-task decoder.
Different from our previous ICCV version \cite{Liu_2021_ICCV}, we deploy SIA instead of self-attention, which selects the foreground regions, splits them into fine-grained segments and aggregates background information into one token. This strategy reduces computational costs by 25\% without significant degradation in results. Furthermore, we incorporate a token-supervised prediction loss in addition to the previous dense prediction scheme to enhance the learning capability of saliency and boundary tokens. For RGB-D SOD, we also propose a depth position embedding method to incorporate 3D depth cues in the decoder with computational efficiency.

\vspace{3mm}
\subsubsection{Token Upsampling and Multi-level Token Fusion}
Due to the relatively small length of $\bm{T}^{\mathcal{C}}$, \ie, $l_3 = h/16 \times w/16$, we contend that generating saliency maps directly from $\bm{T}^{\mathcal{C}}$ tends to limit dense prediction performance and result in low-quality results.
Hence, we adopt a two-step approach: first, we upsample patch tokens and then integrate them with encoder features to facilitate dense prediction.
Rather than employing the commonly used bilinear upsampling approach found in most CNN-based methods \cite{GateNet,zhao2019EGNet,liu2020S2MA,Fu2020JLDCF} for recovering large-scale feature maps, we present a new token upsampling method within the pure transformer framework. Taking inspiration from the T2T module \cite{yuan2021tokens}, which gradually decreases token length by combining neighboring tokens, we propose a reverse T2T (RT2T) transformation to upsample tokens by enlarging each token into multiple sub-tokens, as shown in Figure~\ref{T2T}(b).

To begin with, the input patch tokens are firstly projected to reduce their embedding dimension from $d=384$ to $c=64$ and then a linear projection is used to expand the embedding dimension from $c$ to $c k^2$.
Every token can be regarded as a $k \times k$ image patch with $s$ overlapping neighboring patches, mimicking the soft split step in T2T. This allows us to reconstruct an image by folding the tokens using $p$ zero-padding. The output image size can be computed in reverse using \eqref{softsplit}. Specifically, given the length of the input patch tokens as $h_{i+1}\times w_{i+1}$, the spatial size of the output image is $h_{i}\times w_{i}$. Finally, we reshape the image back to obtain the upsampled tokens with size $l_{i} \times c$, where $l_{i}=h_{i}\times w_{i}$.

By setting $s<k-1$, the RT2T transformation facilitates an increase in token length. Inspired by T2T-ViT, we employ RT2T three times with parameter settings of $k = [3, 3, 7]$, $s = [1, 1, 3]$ and $p = [1, 1, 3]$. As a result, the length of the patch tokens can progressively be upsampled to match $h\times w$, which corresponds to the original input image size.

Besides, drawing inspiration from the demonstrated achievements of multi-level feature fusion in existing SOD methods \cite{hou2018dss,MINet-CVPR2020,GateNet,fan2020bbsnet,luo2020Cas-Gnn}, we make use of low-level tokens with larger lengths from the encoder to provide precise local structural details.
For both RGB and RGB-D SOD, we exclusively use low-level tokens derived from the RGB transformer encoder. 
{Specifically, we gradually fuse $\bm{T}_{i}$ with the upsampled patch tokens $\bm{T}^{\mathcal{D}}_{i+1}$through concatenation and linear projection, formulated as:}
\begin{equation} \label{decoder}
\bm{T}^{\mathcal{D}}_{i} = \text{Linear}([\text{RT2T}(\bm{T}^{\mathcal{D}}_{i+1}),\bm{T}_{i}]),
\end{equation}  
where $i=2,1$ and $[,]$ represents concatenation along the token embedding dimension. ``Linear" denotes a linear projection that decreases the embedding dimension to $c$ after the concatenation. Afterward,
another linear projection is used to restore the embedding
dimension of $\bm{T}^{\mathcal{D}}_{i}$ back to $d$.

\vspace{3mm}
\subsubsection{Token Based Multi-task Prediction}
Previous pure transformer methods \cite{yuan2021tokens,dosovitskiy2020image} commonly utilize a learnable class token for image classification.
Building on this idea, we also incorporate task-related tokens to conduct predictions. Nevertheless, \cite{yuan2021tokens,dosovitskiy2020image} used MLPs on class tokens to generate classification probabilities, which cannot be directly applied for dense prediction in SOD. Thus, we propose to conduct patch-task-attention to aggregate task-related knowledge and perform SOD prediction. 

Besides, drawing inspiration from recent advancement in introducing boundary detection in SOD models \cite{zhao2019EGNet,CVPR2020_LDF,zhang2020select,Wei2020CoNet}, we also follow the multi-task learning scheme to introduce the boundary detection task to saliency prediction, promoting the performance of saliency prediction by facilitating the exchange between object and boundary information.

To this end, we design two task-related tokens: a saliency token $\bm{t}^s\in{\mathbb{R}^{1 \times d}}$ and a boundary token $\bm{t}^b\in{\mathbb{R}^{1 \times d}}$.
We concatenate them with the patch token sequence $\bm{T}^{\mathcal{D}}_{i}$ and process them using $L^{\mathcal{D}}_i$ transformer layers. In each layer, the input two task tokens are used as the output from the previous layer, \ie:
\begin{equation} \label{self-attention}
\begin{bmatrix}
\bm{t}^s_i \\
\bm{T}^{\mathcal{D}}_{i} \\
\bm{t}^b_i
\end{bmatrix}
\leftarrow\text{MLP}(\text{MSA}(
\begin{bmatrix}
\bm{t}^s_{i+1} \\
\bm{T}^{\mathcal{D}}_{i} \\
\bm{t}^b_{i+1}
\end{bmatrix}
)).
\end{equation}
In each layer, self-attention is used to interact the two task tokens with patch tokens. This way enables the aggregation of task-related information from image patches to the two task tokens and also propagates task-specific information from the task tokens to the patch tokens. 
After that, we upsample the enhanced patch tokens to the $i-1$ level and fuse the corresponding level of encoder patch tokens as in \eqref{decoder}, obtaining the patch tokens $\bm{T}^{\mathcal{D}}_{i-1}$. The updated task tokens are then reused to further update themselves and the $\bm{T}^{\mathcal{D}}_{i-1}$ in the next layer. This process is repeated until we reach the final decoder level at the ${1}/{4}$ scale.

Based on the two task-related tokens,
we conduct dense prediction to concurrently generate saliency and boundary predictions.
Specifically, we start by embedding $\bm{T}^{\mathcal{D}}_{i}$ into queries $\bm{Q}^{\mathcal{D}}_{s_{i}} \in {\mathbb{R}^{l_i\times d}}$, $\bm{t}^s_i$ into a key $\bm{K}_{s_{i}} \in {\mathbb{R}^{1\times d}}$ and a value $\bm{V}_{s_{i}} \in {\mathbb{R}^{1\times d}}$ for saliency prediction.
Similarly, for boundary prediction, we embed $\bm{T}^{\mathcal{D}}_{i}$ into $\bm{Q}^{\mathcal{D}}_{b_{i}}$, $\bm{t}^b_i$ into $\bm{K}_{b_{i}}$ and $\bm{V}_{b_{i}}$.
Then, we use the patch-task-attention to obtain the task-related patch tokens:
\begin{equation} \label{predict_patch_task_attention}
\begin{split}
\bm{T}_{s_{i}}^{\mathcal{D}} = \text{sigmoid}(\bm{Q}_{s_{i}}^{\mathcal{D}}\bm{K}_{s_{i}}^{\top} / \sqrt{d})\bm{V}_{s_{i}} + \bm{T}^{\mathcal{D}}_{i}, \\
\bm{T}_{b_{i}}^{\mathcal{D}} = \text{sigmoid}(\bm{Q}_{b_{i}}^{\mathcal{D}}\bm{K}_{b_{i}}^{\top} / \sqrt{d})\bm{V}_{b_{i}} + \bm{T}^{\mathcal{D}}_{i}.
\end{split}
\end{equation}
Since we only have a single key in each equation, here we use the Sigmoid activation for attention computation.

Subsequently, we apply two linear transformations with the Sigmoid activation to map $\bm{T}_{s_{i}}^{\mathcal{D}}$ and $\bm{T}_{b_{i}}^{\mathcal{D}}$ into single channel within the range of $[0,1]$. The results are then reshaped into a 2D saliency map $\bm{y}^{p}_{s_{i}}$ and a 2D boundary map $\bm{y}^{p}_{b_{i}}$, respectively. We perform such prediction at each decoder layer, until reaching the full resolution level. At this level, we upsample the patch tokens of $1/4$ size via the RT2T transformation to obtain the full resolution patch tokens and apply the same dense prediction method.

\begin{figure*}[!t]
  \graphicspath{{Figures/Network/}}
  \centering
  \includegraphics[width=1\linewidth]{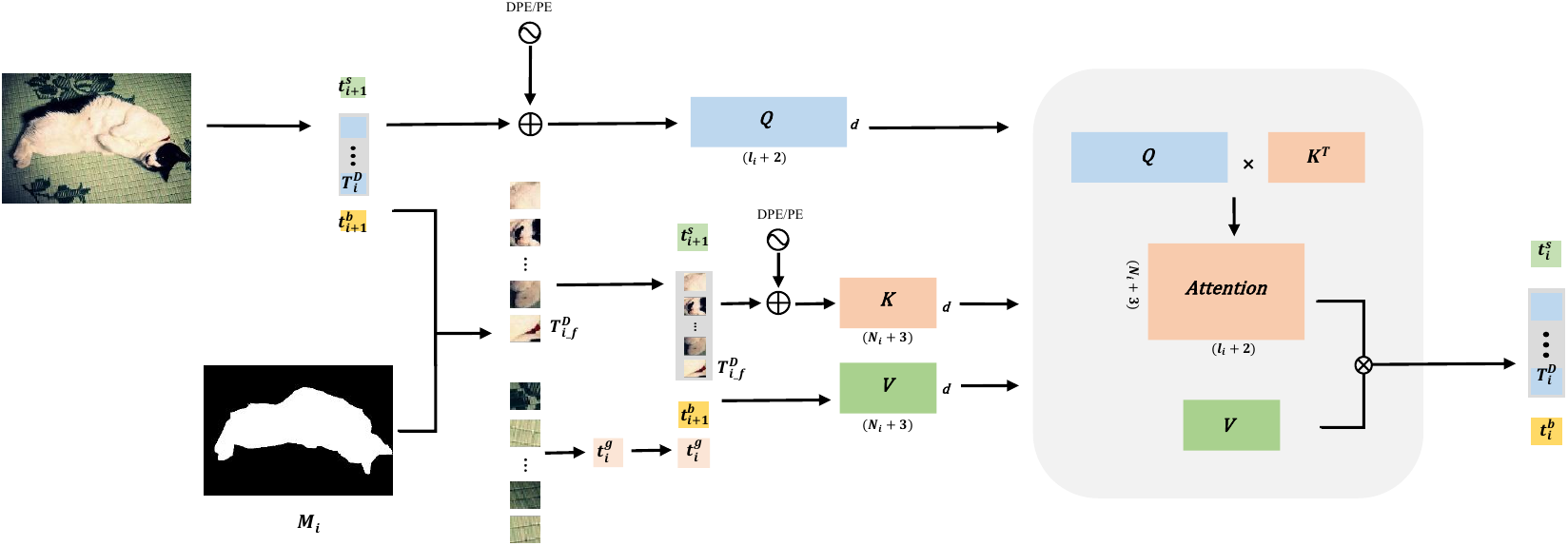}
  \caption{\textbf{Architecture of the SIA module.} We utilize the upsampled and binarized mask $\bm{M}_i$ from the previous stage to select the foreground patch tokens as $\bm{T}^{\mathcal{D}}_{i\_f}$. The background regions are then integrated into a background token $t_i^g$. We use them to replace the original patch tokens $\bm{T}^{\mathcal{D}}_{i}$ to generate key and value, while $\bm{T}^{\mathcal{D}}_{i}$ is used to obtain query for performing cross attention. The two task tokens are also used in the query, key, and value. Depth Position Embedding (DPE) or sinusoidal position encoding (PE) is added to the query and key of the SIA for RGB-D SOD and RGB SOD, respectively.
  }
 
  \label{fig:sia}
\end{figure*}

\vspace{3mm}
\subsubsection{Select-Integrate Attention}
Due to the quadratic computational costs associated with self-attention, our previous VST model \cite{Liu_2021_ICCV} faced significant computational challenge. Taking inspiration from recent works that combine coarse-grained global attention and fine-grained local attention approaches  \cite{liu2021Swin, vaswani2021scaling, zhang2021multi, beltagy2020longformer, yang2021focal} to optimize computational costs, we design Select-Integrate Attention (SIA) specifically aimed at reducing computational burden in the decoder layers.

In the context of the SOD task, foreground information plays a pivotal role, and the background information provides essential contextual cues. To strike a balance between foreground and background, we aggregate background information into a single token, representing the coarse-grained segments, while simultaneously partitioning the foreground regions into fine-grained segments.

Specifically, we use SIA to replace the original self-attention in the decoder. In layer $i$, we first introduce the saliency map generated from the previous stage, \ie, $\bm{y}^{p}_{s_{i+1}}$.
Then, we upsample it twice and binarize it using a threshold of 0.5 for processing it as a mask $\bm{M}_{i}\in{\mathbb{R}^{1 \times l_{i}}}$.

For fine-grained segments, we partition the foreground regions via selecting patches from $\bm{T}^{\mathcal{D}}_{i}$ according to the mask: 
\begin{equation} \label{select-token}
\bm{T}^{\mathcal{D}}_{i\_f} = [\bm{T}^{\mathcal{D}}_{i}(x, y) \mid M_{i}(x, y) = 1].
\end{equation}

For coarse-grained segments, the background patches, which correspond to the index with a value of 0 in the mask $\bm{M}_{i}$, are integrated into a background token $\bm{t}^g_i \in {\mathbb{R}^{1 \times d}}$ via average pooling: 
\begin{equation} \label{backgroundtokens}
\bm{t}^g_i = \dfrac{{(1-\bm{M}_{i})}{\bm{T}^{\mathcal{D}}_{i}}}{\begin{matrix} \sum_{k=1}^{l_{i}} (1-{\bm{M}_{i}[:,k]}) \end{matrix}}.
\end{equation}
Afterwards, we use the concatenation of $\bm{T}^{\mathcal{D}}_{i}$ and the two task tokens to generate query, and use the concatenation of $\bm{T}^{\mathcal{D}}_{i\_f}$, $\bm{t}^g_i$, and the task tokens to generate key and value to perform cross attention:
\begin{equation} \label{sia}
\begin{bmatrix}
\bm{t}^s_i \\
\bm{T}^{\mathcal{D}}_{i} \\
\bm{t}^b_i
\end{bmatrix}
\leftarrow\text{MLP}(\text{MCA}(
\begin{bmatrix}
\bm{t}^s_{i+1} \\
\bm{T}^{\mathcal{D}}_{i} \\
\bm{t}^b_{i+1}
\end{bmatrix}
,
\begin{bmatrix}
\bm{t}^s_{i+1} \\
\bm{T}^{\mathcal{D}}_{i\_f} \\
\bm{t}^g_i \\
\bm{t}^b_{i+1}
\end{bmatrix})),
\end{equation}
where MCA means using the standard multi-head attention \cite{vaswani2017attention} in the cross attention (CA) \eqref{cross-at}. The structure of our proposed SIA is shown in Figure~\ref{fig:sia}.

Above operation enables us to construct our SIA by considering both fine-grained and coarse-grained context propagation.
As there is no mask comes from the previous stage at the first 1/16 decoder level, we retain the self-attention used in \eqref{self-attention} for this decoder level and apply SIA in the 1/8 and 1/4 decoder levels.

Assuming decoder layer $i$ contains $l_{i}$ patch tokens, in which there are only $N_{i}$ foreground tokens, the computation complexity of each self-attention is $\bm{O}((l_{i} +2)^{2})$,
while the computation complexity of SIA becomes $\bm{O}((N_{i}+3)(l_{i}+2))$. Since $(N_{i}+1)<l_{i}$, this approach ensures saved computational costs while preserving accurate information in foreground regions and also retaining global background context.

Since our SIA breaks the original spatial structure of the patch tokens, we adopt the sinusoidal position encoding introduced in \cite{vaswani2017attention} for $\bm{T}^{\mathcal{D}}_{i\_f}$ and $\bm{T}^{\mathcal{D}}_{i}$. Additionally, we employ three learnable position encodings for the saliency token, boundary token, and background token. 

It is essential to acknowledge that the training method of our SIA differs from that of testing. Due to the uncertainty regarding the number of selected foreground patches, which hinders parallel computation, we still use all patch tokens and adopt the masked attention \cite{cheng2022masked} to filter out background patch tokens during training.

\subsubsection{Depth Position Embedding}
The sinusoidal position encoding in SIA incorporates 2D position information. Nevertheless, for RGB-D SOD, we are also provided the 3D depth structure, which was ignored in our previous VST model \cite{Liu_2021_ICCV}.
Therefore, we design a novel depth position encoding (DPE) method building upon the 2D sinusoidal position encoding. In this way, we also efficiently integrate the depth information in the decoder and do not alter the overall architecture and introduce much computational cost as previous two-branch RGB-D SOD methods did.

The main challenge we aim to address is the difference in nature between depth as a continuous variable and 2D coordinates as discrete variables. To tackle this, for each decoder level $i$, we first obtain a resized depth map from the initial depth map to match the spatial dimension of $h_i \times w_i$.
Next, we normalize the depth map to [0,1] and then multiply the normalized depth values with $h_i$ and round them up to obtain a discrete depth map. This makes the depth values have the same value range as the 2D coordinates.
The later step is similar to sinusoidal position encoding, which is formulated as:
\begin{equation} \label{depthPE}
\begin{split}
\bm{DPE}_{(dep,2m)} = \sin{(dep/10000^{2m/d_{model}})}, \\
\bm{DPE}_{(dep,2m+1)} = \cos{(dep/10000^{2m/d_{model}})},
\end{split}
\end{equation}
where $dep$ represents the depth value and $m$ denotes the dimension. In addition, we propose three learnable scaling factors, namely $Z_i$, for the three decoder layers, to establish the relative importance of DPEs compared to 2D spatial PEs. We multiply each scaling factor with the corresponding DPE and then concatenate it with the 2D spatial PEs to obtain 3D PEs. The 3D PEs are added to the query and key of the SIA or the self-attention in the decoder layers to provide 3D structural priors.

\subsection{Loss Function}
In our previous VST model \cite{Liu_2021_ICCV}, we only adopted BCE losses on the saliency and boundary predictions, \ie, $\bm{y}^{p}_{s_{i}}$ and $\bm{y}^{p}_{b_{i}}$, at each decoder level.
However, we have observed that the predictions are directly obtained from the enhanced patch tokens, \ie, $\bm{T}^{\mathcal{D}}_{s_i}$ and $\bm{T}^{\mathcal{D}}_{b_i}$, with two linear transformations.
Consequently, the dense predictions are not directly related to the two task tokens, hence may hinder their ability of effectively learning task-related information.

Hence, in this paper, we propose a token-supervised prediction loss, directly providing supervision for saliency and boundary tokens. Specifically, in each decoder layer $i$, we directly perform inner product between the two task tokens and the patch tokens, thus generating two segmentation predictions $\bm{y}^{s}_{s_{i}}$ and $\bm{y}^{s}_{b_{i}}$, formulated as follows:
\begin{equation} \label{predict_segmentation_task}
\begin{split}
\bm{y}^{s}_{s_{i}} = \bm{t}^s_i {\bm{T}^{\mathcal{D}}_{i}}^{\top} / \sqrt{d}, \\
\bm{y}^{s}_{b_{i}} = \bm{t}^b_i {\bm{T}^{\mathcal{D}}_{i}}^{\top} / \sqrt{d}.
\end{split}
\end{equation} 
Next, we use the Sigmoid activation and reshape them to 2D saliency and boundary maps for deploying BCE losses. In this way, the gradient from the loss can directly flow to the task-related tokens $\bm{t}^s_i$ and $\bm{t}^b_i$, hence enhancing their learning efficiency.

In summary, we adopt the dense prediction loss $\mathcal{L}_{den}^{i}$ and the token-supervised prediction loss $\mathcal{L}_{sup}^{i}$ at levels $1$, $1/4$, $1/8$, and $1/16$ to optimize each decoder stage, as follows:
\begin{equation}\label{loss}
\begin{split}
\mathcal{L}_s = \sum_{i=1}^{4}(\mathcal{L}_{den}^{i}(\widehat{\bm{y}_{s_{i}}},\bm{y}^{p}_{s_{i}}) + \mathcal{L}_{sup}^{i}(\widehat{\bm{y}_{s_{i}}},\bm{y}^{s}_{s_{i}})), \\
\mathcal{L}_b = \sum_{i=1}^{4}(\mathcal{L}_{den}^{i}(\widehat{\bm{y}_{b_{i}}},\bm{y}^{p}_{b_{i}}) + \mathcal{L}_{sup}^{i}(\widehat{\bm{y}_{b_{i}}},\bm{y}^{s}_{b_{i}})),
\end{split}
\end{equation}
where $\mathcal{L}_s$ and $\mathcal{L}_b$ represent the saliency prediction loss and the boundary prediction loss in total, respectively. $\widehat{\bm{y}_{s_{i}}}$ and $\widehat{\bm{y}_{b_{i}}}$ denote the ground truth for saliency and boundary tasks, respectively. The $\mathcal{L}_{den}^{i}$ and $\mathcal{L}_{sup}^{i}$ are implemented as BCE losses.
The ultimate loss function is formulated as their summation:
\begin{equation}\label{total_loss}
\mathcal{L}_{total} = \mathcal{L}_s + \mathcal{L}_b.
\end{equation}

\section{Experiments}
\subsection{Datasets and Evaluation Metrics}
For RGB SOD, we evaluate our proposed models using six commonly used benchmark datasets, which are introduced next. \textbf{DUTS} \cite{wang2017duts} is currently the largest available dataset for salient object detection, with 10,553 training images and 5,019 testing images.  \textbf{ECSSD} \cite{yan2013ECSSD} includes 1,000 semantically meaningful images, while \textbf{HKU-IS} \cite{li2015HKUIS} consists of 4,447 images with multiple foreground objects. \textbf{PASCAL-S} \cite{li2014PASCALS} contains 850 images collected from the PASCAL VOC 2010 dataset \cite{everingham2010pascal}. 
\textbf{DUT-O} \cite{yang2013DUTO} includes 5,168 images and \textbf{SOD} \cite{movahedi2010SOD} contains 300 images, respectively.

For RGB-D SOD, nine widely used benchmark datasets are used for evaluation. \textbf{STERE} \cite{niu2012stere}, the first stereoscopic saliency dataset, contains 1,000 images. \textbf{NJUD} \cite{ju2014njud} includes 1,985 images collected from the Internet, 3D movies, and photographs. 
\textbf{NLPR} \cite{peng2014nlpr}, \textbf{DUTLF-Depth} \cite{Piao2019dmra}, and \textbf{SIP} \cite{fan2020SIP} are captured using Microsoft Kinect, a light field camera, and a Huawei Mate 10 smartphone, respectively. They consist of 1,000 images, 1,200 images, and 929 salient person images, respectively.  \textbf{LFSD} \cite{li2014lfsd}, \textbf{RGBD135} \cite{cheng2014rgbd135}, and \textbf{SSD} \cite{zhu2017ssd} are three small-scale datasets with 100, 135, and 80 images, respectively. \textbf{ReDWeb-S} \cite{liu2021learning} contains 3,179 images with diverse and challenging visual scenes and high-quality depth maps.

Following recent works, we employ four widely used evaluation metrics to assess model performance. Structure-measure $S_m$ \cite{fan2017structure} takes into account structural similarity both at region-level and object-level. Maximum F-measure (maxF) treats SOD as a binary classification task and considers precision and recall simultaneously under varying thresholds. Finally, it reports the highest score under the optimal threshold. 
Maximum enhanced-alignment measure $E_{\xi}^{\text{max}}$ \cite{Fan2018Enhanced} combines local pixel values in an item with image-level averages to capture image-level statistics and local-pixel matching information. 
The last metric we use is Mean Absolute Error (MAE), which computes pixel-wise average absolute error. 
Meanwhile, we report the number of parameters (Params) and the multiply accumulate operations (MACs) to better evaluate the model's computational complexity.

\begin{table*}[t]
\centering
\scriptsize
\renewcommand{\arraystretch}{1.0}
\renewcommand{\tabcolsep}{1.0mm}
\caption{\textbf{Ablation study of our extended model designs for RGB SOD.}
We use our previous model VST \cite{Liu_2021_ICCV} as baseline and show the results and MACs of gradually using $\mathcal{L}_{sup}$ (token-supervised prediction loss) and SIA (select-integrate attention). The best results are labeled in \blu{blue}.}
\resizebox{\linewidth}{!}{
\begin{tabular}{ccc|c|cccc|cccc|cccc}
\hline
 \multicolumn{3}{c|}{Settings} & {\multirow{2}{*}{MACs(G)}} & \multicolumn{4}{c|}{ECSSD\cite{yan2013ECSSD}} & \multicolumn{4}{c|}{DUT-O\cite{yang2013DUTO}} & \multicolumn{4}{c}{SOD\cite{movahedi2010SOD}} \\ 
 \multicolumn{1}{c}{VST}  & \multicolumn{1}{c}{$\mathcal{L}_{sup}$} & \multicolumn{1}{c|}{SIA}  &{} &\multicolumn{1}{l}{$S_m \uparrow$} & \multicolumn{1}{l}{maxF $\uparrow$} & \multicolumn{1}{l}{$E_{\xi}^{\text{max}} \uparrow$} & \multicolumn{1}{l|}{MAE $\downarrow$} 
                      & \multicolumn{1}{l}{$S_m \uparrow$} & \multicolumn{1}{l}{maxF $\uparrow$} & \multicolumn{1}{l}{$E_{\xi}^{\text{max}} \uparrow$} & \multicolumn{1}{l|}{MAE $\downarrow$}
                      & \multicolumn{1}{l}{$S_m \uparrow$} & \multicolumn{1}{l}{maxF $\uparrow$} & \multicolumn{1}{l}{$E_{\xi}^{\text{max}} \uparrow$} & \multicolumn{1}{l}{MAE $\downarrow$}
  \\ \hline
  
{\checkmark} &{} &{} &{41.36\footnotemark[1]} &0.932 &0.944 &0.964 &0.034   &0.850 &0.800 &0.888 &0.058   &0.854 &0.866 &0.902 &0.065  \\
{\checkmark} &{\checkmark} &{} &{42.67} &\blu{0.935} &\blu{0.948} &\blu{0.968} &\blu{0.029}   &\blu{0.854} &0.810 &\blu{0.894} &\blu{0.055}   &\blu{0.856}  &0.868 &\blu{0.905} &\blu{0.061}  \\
{\checkmark} &{\checkmark} &{\checkmark} &{30.07} &0.933 &0.946 &0.966 &0.030   &\blu{0.854} &\blu{0.812} &\blu{0.894} &0.056   &0.855 &\blu{0.870} &\blu{0.905} &0.062 \\\hline

\end{tabular}}
\label{ablationTabRGB}
\end{table*}

\begin{table*}[t]
\centering
\scriptsize
\renewcommand{\arraystretch}{1.0}
\renewcommand{\tabcolsep}{1.0mm}
\caption{
\textbf{Ablation study of our extended model designs for RGB-D SOD.} We use our previous model VST \cite{Liu_2021_ICCV} as baseline and show the results  and MACs of gradually using $\mathcal{L}_{sup}$ (token-supervised prediction loss), SIA (select-integrate attention), and DPE (depth position encoding). The best results are labeled in \blu{blue}.
}
\resizebox{\linewidth}{!}{
\begin{tabular}{cccc|c|cccc|cccc|cccc}
\hline
 \multicolumn{4}{c|}{Settings} & {\multirow{2}{*}{MACs(G)}} &\multicolumn{4}{c|}{NLPR\cite{peng2014nlpr}} & \multicolumn{4}{c|}{ReDWeb-S\cite{liu2021learning}} & \multicolumn{4}{c}{SSD\cite{zhu2017ssd}} \\
  {VST} &{$\mathcal{L}_{sup}$} & {SIA} &{DPE} &{} &\multicolumn{1}{l}{$S_m \uparrow$} & \multicolumn{1}{l}{maxF $\uparrow$} & \multicolumn{1}{l}{$E_{\xi}^{\text{max}} \uparrow$} & \multicolumn{1}{l|}{MAE $\downarrow$} 
                      & \multicolumn{1}{l}{$S_m \uparrow$} & \multicolumn{1}{l}{maxF $\uparrow$} & \multicolumn{1}{l}{$E_{\xi}^{\text{max}} \uparrow$} & \multicolumn{1}{l|}{MAE $\downarrow$}
                      & \multicolumn{1}{l}{$S_m \uparrow$} & \multicolumn{1}{l}{maxF $\uparrow$} & \multicolumn{1}{l}{$E_{\xi}^{\text{max}} \uparrow$} & \multicolumn{1}{l}{MAE $\downarrow$}
  \\ \hline
 {\checkmark} &{} &{} &{} &51.33\footnotemark[1] &0.932 &0.920 &0.962 &0.024   &0.759 &0.763 &0.826 &0.113   &0.889 &0.876 &0.935 &0.045  \\ 
 {\checkmark} &{\checkmark} &{} &{} &52.64 &\blu{0.934} &\blu{0.923} &0.964 &0.022   &0.763 &0.768 &0.831 &0.108   &\blu{0.896} &\blu{0.884} &0.940 &\blu{0.038} \\ 
 {\checkmark} &{\checkmark} &{\checkmark} &{} &39.98 &0.931 &0.922 &0.963 &0.023  &0.766 &0.771 &0.830 &0.108  &0.891 &0.873 &0.940 &0.040\\
 {\checkmark} &{\checkmark} &{\checkmark} &{\checkmark} &39.95 &\blu{0.934} &0.922 &\blu{0.966} &\blu{0.020}   &\blu{0.767} &\blu{0.771} &\blu{0.835} &\blu{0.103}   &\blu{0.896} &0.883 &\blu{0.944} &\blu{0.038}\\\hline

\end{tabular}}
\label{ablationTabRGBD}
\end{table*}
 
\vspace{1mm}
\subsection{Implementation Details}
Previous SOD methods \cite{GateNet,gao2020sod100k,CVPR2020_LDF,MINet-CVPR2020} commonly employ the DUTS training set to train their RGB SOD models. Likewise, we train our network using the same dataset. For RGB-D SOD, we follow \cite{zhang2020select, piao2020a2dele, Wei2020CoNet, zhang2020ATSA} to construct our training set, which includes 1,485 images from NJUD, 700 images from NLPR, and 800 images from DUTLF-Depth.

To generate the boundary ground truth from GT saliency maps, we apply the sober operator as described in \cite{zhao2019EGNet}.
We conduct all experiments using the publicly available Pytorch library \cite{paszke2019pytorch}. Training and testing are performed on a GTX 1080 Ti GPU. For data preprocessing, the original single-channel depth maps are normalized to the range of [0,1] and duplicated to three channels. Both RGB and depth images are resized to $256 \times 256$ pixels and then randomly cropped to $224 \times 224$ image regions for training. Data augmentation techniques, such as random flipping, are also applied.
In our convertor and decoder, we set $L^{\mathcal{C}}=L^{\mathcal{D}}_3=4$ and $L^{\mathcal{D}}_2=L^{\mathcal{D}}_1=2$ according to experimental results. 
For RGB SOD, the batchsize is set to 9 (T2T-ViT$_t$-14 /Swin-T /Swin-S ) or 8 (Swin-B), while for RGB-D SOD it is set to 7 (T2T-ViT$_t$-14 /Swin-T /Swin-S ) or 6 (Swin-B). The total training steps are 60,000 and 40,000 for RGB and RGB-D, respectively. We adopt the Adam optimizer with an initial learning rate of 0.0001, which is reduced by a factor of 10 at half and three-quarters of the total training step.

\footnotetext[1]{Please note that here the number of MACs is different from that in our original conference paper \cite{Liu_2021_ICCV}. This is because the toolbox we used to compute the MACs for our conference paper did not count in matrix multiplication operations in attention modules of Transformers. In this paper, we updated the toolbox and fixed the bug, hence obtaining accurate MACs.}

\subsection{Ablation Study}\label{sec:ablation}
We perform ablation experiments on three commonly used RGB SOD datasets and three RGB-D SOD datasets to showcase the efficacy of our proposed model elements. The experimental results are presented in Table~\ref{ablationTabRGB} and Table~\ref{ablationTabRGBD}. It is important to note that in this context, we exclusively conduct ablation studies for the extended components in VST\texttt{++} and use our previous VST model as the baseline since the components in VST have already proved their effectiveness in our prior VST paper \cite{Liu_2021_ICCV}.

\subsubsection{Effectiveness of Token-supervised Prediction Loss}
Our findings demonstrate that incorporating the token-supervised prediction loss ($\mathcal{L}_{sup}$) can enhance the overall detection performance on most RGB and RGB-D SOD datasets. This notable enhancement can be attributed to the direct learning ability of task-related tokens brought by our token-supervised prediction loss.

\begin{table*}[t]
  \centering
  \footnotesize
  \renewcommand{\arraystretch}{1.0}
  \renewcommand{\tabcolsep}{1mm}
 \caption{\textbf{{Quantitative comparison of our proposed VST\texttt{++} with other 14 SOTA RGB SOD methods on six benchmark datasets.}} Here we employ the average MACs value of our model on all datasets since the number of the selected tokens is different for each image. 
 ``-R",``-R2", ``-t", ``-T", ``-B", and ``-S"  mean the ResNet50 \cite{he2016resnet}, Res2Net \cite{gao2019res2net}, T2T-ViT$_t$-14 \cite{yuan2021tokens}, SwinT-1k, SwinS-1k, and SwinB-22k \cite{liu2021Swin} backbones, respectively. In our VST-B\texttt{++}, we use a larger input image resolution ($228\times 228$) to ensure comparable computational costs with ICON-B, which uses a $384\times 384$ input resolution. For our remaining VST and VST\texttt{++} counterparts, we consistently use a $224\times 224$ input resolution.
 \red{Red} and \blu{blue} denote the best and the second-best results, respectively.
 }
 \begin{adjustbox}{width=\linewidth}
 \begin{tabular}{l|c|c|cccc|cccc|cccc|cccc|cccc|cccc}
  \hline

  \multicolumn{3}{c|}{Summary} & \multicolumn{4}{c|}{DUTS\cite{wang2017duts}} & \multicolumn{4}{c|}{ECSSD\cite{yan2013ECSSD}} & \multicolumn{4}{c|}{HKU-IS\cite{li2015HKUIS}} & \multicolumn{4}{c|}{PASCAL-S\cite{li2014PASCALS}} & \multicolumn{4}{c|}{DUT-O\cite{yang2013DUTO}} & \multicolumn{4}{c}{SOD\cite{movahedi2010SOD}}\\
  \hline
   \multicolumn{1}{l|}{Metric} & \multicolumn{1}{c|}{MACs} &\multicolumn{1}{c|}{Params}
    & $S_m$ & maxF & $E_{\xi}^{\text{max}}$ & MAE & $S_m$ & maxF & $E_{\xi}^{\text{max}}$ & MAE & $S_m$ & maxF & $E_{\xi}^{\text{max}}$ & MAE & $S_m$ & maxF & $E_{\xi}^{\text{max}}$ & MAE & $S_m$ & maxF & $E_{\xi}^{\text{max}}$ & MAE & $S_m$ & maxF & $E_{\xi}^{\text{max}}$ & MAE\\
   \hline
   \multicolumn{27}{c}{CNN-based} \\\hline
   PiCANet\cite{liu2018picanet} &54.05 &47.22 &0.863 &0.840 &0.915 &0.040 &0.916 &0.929 &0.953 &0.035 &0.905 &0.913 &0.951 &0.031 &0.846 &0.824 &0.882 &0.071 &0.826 &0.767 &0.865 &0.054 &0.813 &0.824&0.871 &0.073\\
   AFNet\cite{Feng_AFNet} &21.66 &35.95 &0.867 &0.838 &0.910 &0.045 &0.914 &0.924&0.947 &0.042 &0.905 &0.910 &0.949 &0.036 &0.849 &0.824 &0.877 &0.076 &0.826 &0.759 &0.861 &0.057 &0.811 &0.819 &0.867 &0.085\\
   TSPOANet\cite{Liu_TSPOANet} &- &- &0.860 &0.828 &0.907 &0.049 &0.907 &0.919 &0.942 &0.047 &0.902 &0.909 &0.950 &0.039 &0.841 &0.817 &0.871 &0.082 &0.818 &0.750 &0.858 &0.062 &0.802 &0.809 &0.852 &0.904\\
   EGNet-R\cite{zhao2019EGNet} &157.21 &111.64 &0.887 &0.866 &0.926 &0.039 &0.925 &0.936 &0.955 &0.037 &0.918 &0.923 &0.956 &0.031 &0.852 &0.825 &0.874 &0.080 &0.841 &0.778 &0.878 &0.053 &0.824 &0.831 &0.875 &0.080\\
   ITSD-R\cite{Zhou2020ITSD} &15.96 &26.47 &0.885 &0.867 &0.929 &0.041 &0.925 &0.939 &0.959 &0.035 &0.917 &0.926 &0.960 &0.031 &0.861 &0.839 &0.889 &0.771 &0.840 &0.792 &0.880 &0.061 &0.835 &0.849 &0.889 &0.075\\
   MINet-R\cite{MINet-CVPR2020} &87.11 &162.38 &0.884 &0.864 &0.926 &0.037 &0.925 &0.938 &0.957 &0.034 &0.919 &0.926 &0.960 &0.029 &0.856 &0.831 &0.883 &0.071 &0.883 &0.769 &0.869 &0.056 &0.830 &0.835&0.878 &0.074\\
   LDF-R\cite{CVPR2020_LDF} &15.51 &25.15 &0.892 &0.877 &0.930 &0.034 &0.925 &0.938 &0.954 &0.034 &0.920 &0.929 &0.958 &0.028 &0.861 &0.839 &0.888 &0.067 &0.839 &0.782 &0.870 &0.052 &0.831 &0.841 &0.878 &0.071\\
   CSF-R2\cite{{gao2020sod100k}} &18.96 &36.53 &0.890 &0.869 &0.929 &0.037 &0.931 &0.942 &0.960 &0.033 &- &- &- &- &0.863 &0.839 &0.885 &0.073 &0.838 &0.775 &0.869 &0.055 &0.826 &0.832 &0.883 &0.079\\
   GateNet-R\cite{GateNet} &162.22 &128.63 &0.891 &0.874 &0.932 &0.038 &0.924 &0.935 &0.955 &0.038 &0.921 &0.926 &0.959 &0.031 &0.863 &0.836 &0.886 &0.071 &0.840 &0.782 &0.878 &0.055 &0.827 &0.835 &0.877 &0.079\\
   {MENet}\cite{wang2023pixels} &94.62 &27.83 &0.905 &0.895 &0.943 &0.028 &0.927 &0.938 &0.956 &0.031 &0.927 &0.939 &0.965 &0.023 &0.871 &0.848 &0.892 &0.062 &0.850 &0.792 &0.879 &0.045 &0.841 &0.847 &0.884 &0.065\\\hline
   \multicolumn{27}{c}{{Generic Dense Prediction Model}} \\\hline
   {Mask2Former-T} \cite{cheng2022masked} &248.16 &47.40 &0.880 &0.871 &0.925 &0.041 &0.922 &0.942 &0.961 &0.037 &0.916 &0.937 &0.967 &0.032 &0.866 &0.845 &0.890 &0.069 &0.823 &0.763 &0.858 &0.062 &0.815 &0.847 &0.882 &0.081\\\hline
   \multicolumn{27}{c}{Transformer-based} \\\hline
    VST\cite{Liu_2021_ICCV} &41.36 &44.48 &0.896 &0.877 &0.939 &0.037 &0.932 &0.944 &0.964 &0.034 &0.928 &0.937 &0.968 &0.030 &0.873 &0.850 &0.900 &0.067 &0.850 &0.800 &0.888 &0.058 &0.854 &0.866 &0.902 &0.065\\
   EBMGSOD\cite{jing_ebm_sod21} &53.38 &118.96 &0.909 &0.900 &0.949 &\blu{0.029} &\blu{0.941} &\blu{0.954} &\blu{0.972} &\blu{0.024} &0.930 &0.943 &0.971 &0.023 &0.877 &0.856 &0.899 &0.061 &0.858 &0.817 &\blu{0.900} &0.051 &\blu{0.866} &\blu{0.881} &\red{0.915} &\red{0.056}\\
   ICON-B\cite{zhuge2021salient} &52.59 &94.30 &\blu{0.917} &\blu{0.911} &\red{0.960} &\red{0.024} &\blu{0.941} &\blu{0.954} &0.971 &\blu{0.024} &\blu{0.936} &\blu{0.947} &\blu{0.974} &\blu{0.022} &\blu{0.885} &\blu{0.860} &\blu{0.903} &\blu{0.058} &\blu{0.869} &\red{0.830} &\red{0.906} &\red{0.043}&0.862 &0.878 &0.909 &\blu{0.059}\\\hline
   \multicolumn{27}{c}{Ours} \\\hline
   VST-t\texttt{++} &30.07 &46.05 &0.897 &0.881 &0.939 &0.035 &0.933 &0.946 &0.966 &0.030 &0.928 &0.938 &0.969 &0.026 &0.869 &0.849 &0.899 &0.066 &0.854 &0.812 &0.894 &0.056 &0.855 &0.870 &0.905 &0.062\\
   VST-T\texttt{++} &28.44 &53.60 &0.901 &0.887 &0.943 &0.033 &0.937 &0.949 &0.968 &0.029 &0.930 &0.939 &0.968 &0.026 &0.878 &0.855 &0.901 &0.063 &0.853 &0.804 &0.892 &0.053 &0.853 &0.866 &0.899 &0.065\\
   VST-S\texttt{++} &32.73 &74.90 &0.909 &0.897 &0.947 &0.029 &0.939 &0.951 &0.969 &0.027 &0.932&0.941 &0.969 &0.025 &0.880 &0.859 &0.901 &0.062 &0.859 &0.813 &0.890 &0.050 &0.859 &0.866 &0.898 &0.059\\
    VST-B\texttt{++} &69.30 &112.23 &\red{0.923} &\red{0.915} &\blu{0.957} &\red{0.024} &\red{0.950} &\red{0.962} &\red{0.977} &\red{0.021} &\red{0.941} &\red{0.951} &\red{0.976} &\red{0.021} &\red{0.889} &\red{0.867} &\red{0.907} &\red{0.056} &\red{0.871} &\blu{0.827} &\blu{0.900} &\blu{0.044} &\red{0.872} &\red{0.886} &\blu{0.910} &\red{0.056}\\\hline
   \end{tabular}
   \end{adjustbox}
  \label{RGB_SOTA}
 \end{table*}

 \begin{figure*}[htbp]
  \graphicspath{{Figures/qualitative/}}
  \centering
  \begin{overpic}[width=1\linewidth]{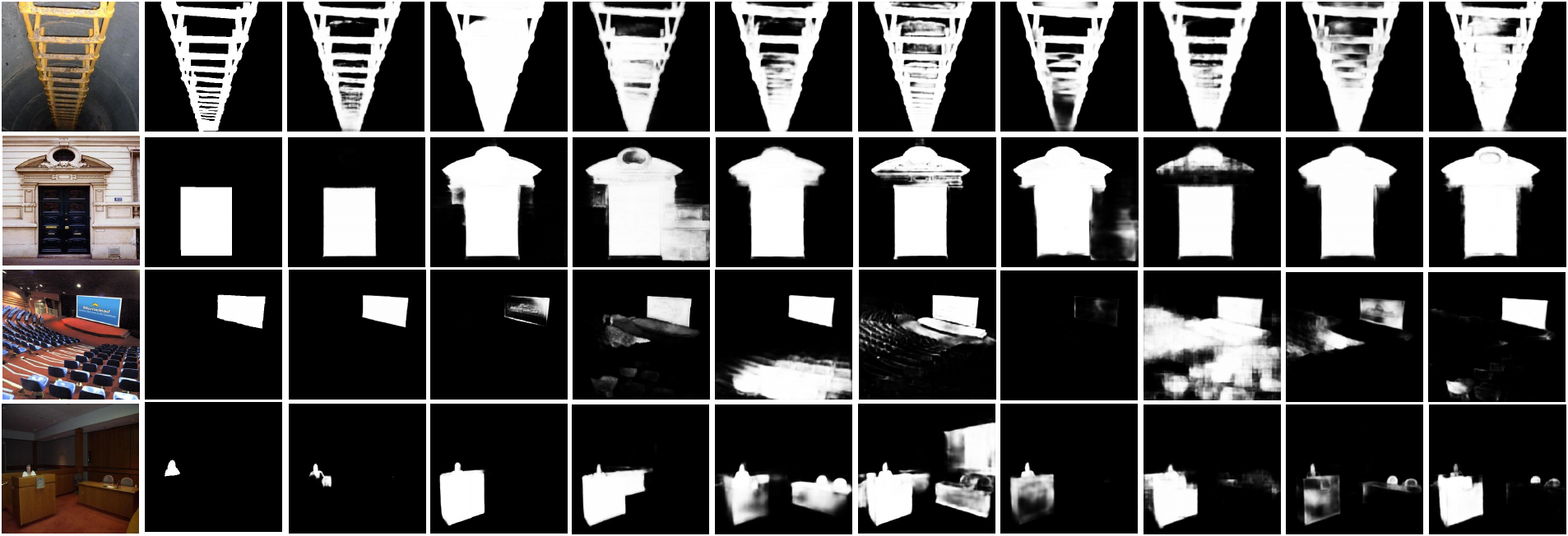}
  \put(2.8,-1.6){\notsotiny {Image}}
  \put(12.5,-1.6){\notsotiny {GT}}
  \put(20.2,-1.6){\notsotiny {VST-B++}}
  \put(29.8,-1.6){\notsotiny {MENet}}
  \put(36.6,-1.6){\notsotiny {Mask2Former}}
  \put(47.8,-1.6){\notsotiny {ICON-B}}
  \put(56.0,-1.6){\notsotiny {EBMGSOD}}
  \put(66.5,-1.6){\notsotiny {LDF-R}}
  \put(76.0,-1.6){\notsotiny {VST}}
  \put(83.4,-1.6){\notsotiny {GateNet-R}}
  \put(93.0,-1.6){\notsotiny {CSF-R2}}
  \end{overpic}
  \caption{\textbf{Qualitative comparison of our model against state-of-the-art RGB SOD methods.} (GT: ground truth.)}
  \label{visualcmpRGB}
\end{figure*}
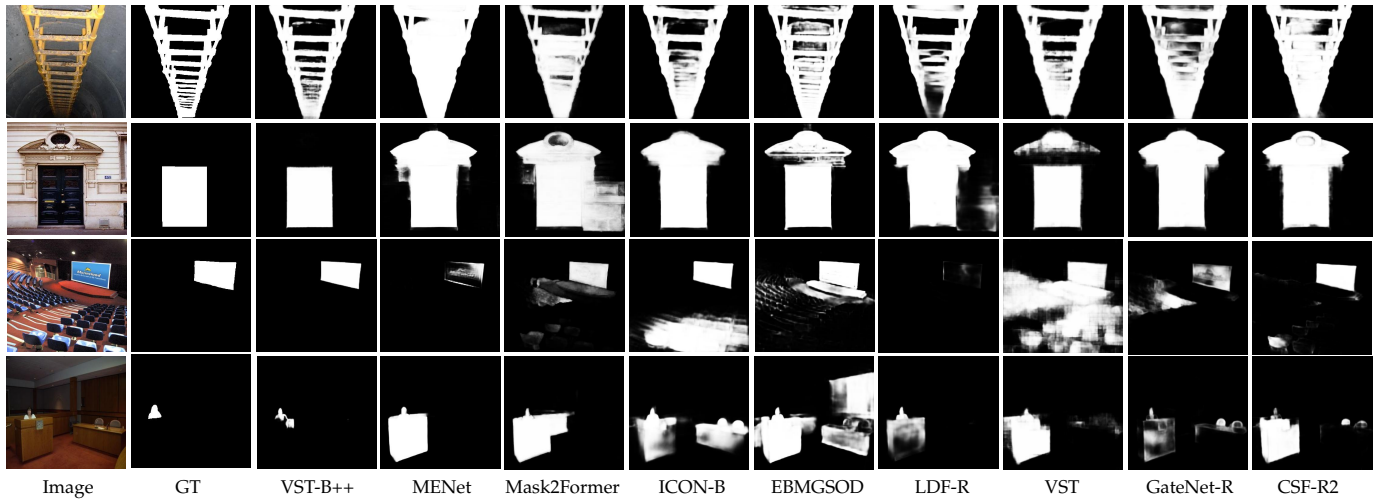

\begin{table*}[t]
  \centering
  \footnotesize
  \renewcommand{\arraystretch}{1.0}
  \renewcommand{\tabcolsep}{0.7mm}
 \caption{\textbf{Quantitative comparison of our proposed VST\texttt{++} with other 19 SOTA RGB-D SOD methods on 4 benchmark datasets.} Here we employ the average MACs value of our model on all RGB-D datasets because the number of the selected tokens is different for each image. `-' indicates the code or result is not available. `-t', `-T', `-S' and `-B' indicate the T2T-ViT$_t$-14 backbone, the SwinT-1k backbone, the SwinS-1k backbone, and the SwinB-22k backbone with $288\times 228$ image size.
 \red{Red} and \blu{blue} denote the best and the second-best results, respectively.}
  \resizebox{\linewidth}{!}{
  \begin{tabular}{l|cc|cccc|cccc|cccc|cccc}
  \hline

  \multicolumn{3}{c|}{Summary} & \multicolumn{4}{c|}{SSD\cite{zhu2017ssd}} & \multicolumn{4}{c|}{RGBD135\cite{cheng2014rgbd135}} & \multicolumn{4}{c|}{LFSD\cite{li2014lfsd}} & \multicolumn{4}{c}{SIP\cite{fan2020SIP}} \\
  \hline
  Metric & MACs(G) & Params(M)
    & $S_m\uparrow$ & maxF$\uparrow$ & $E_{\xi}^{\text{max}}\uparrow$ & MAE$\downarrow$  & $S_m\uparrow$ & maxF$\uparrow$ & $E_{\xi}^{\text{max}}\uparrow$ & MAE$\downarrow$  & $S_m\uparrow$ & maxF$\uparrow$ & $E_{\xi}^{\text{max}}\uparrow$ & MAE$\downarrow$  & $S_m\uparrow$ & maxF$\uparrow$ & $E_{\xi}^{\text{max}}\uparrow$ & MAE$\downarrow$ \\
   \hline
   \multicolumn{19}{c}{CNN-based} \\\hline
   $S^2$MA\cite{liu2020S2MA}  & 141.19 &86.65 & 0.868 &0.848 &0.909 &0.053 &0.941 &0.935 &0.973 &0.021 &0.829 &0.831 &0.965 &0.102 &0.872 &0.877 &0.919 &0.058\\ 
   PGAR\cite{chen2020PGAR} & 44.65 &16.2 &0.832 &0.798 &0.872 &0.068 &0.886 &0.864 &0.924 &0.032 &0.808 &0.794 &0.853 &0.099 &0.838 &0.827 &0.886 &0.073 \\
   DANet\cite{zhao2020DANet} &66.25 &26.68 &0.864 &0.843 &0.914 &0.050 &0.924 &0.914 &0.966 &0.023 &0.841 &0.840 &0.874 &0.087 &0.875 &0.876 &0.918 &0.055 \\
   cmMS\cite{li2020cmMS} &134.77 &92.02 &0.857 &0.839 &0.900 &0.053 &0.934 &0.928 &0.969 &0.018 &0.845 &0.858 &0.886 &0.082 &0.872 &0.876 &0.911 &0.058\\
   ATST\cite{zhang2020ATSA} &42.17 &32.17 &0.850 &0.853 &0.920 &0.052 &0.917 &0.916 &0.961 &0.922 &0.845 &0.859 &0.893 &0.078 &0.849 &0.861 &0.901 &0.063\\
   CMW\cite{Li2020CMWNet} &208.03 &85.65 &0.798 &0.771 &0.871 &0.085 &0.934 &0.931 &0.969 &0.022 &0.776 &0.779 &0.834 &0.130 &0.705 &0.677 &0.804 &0.141 \\
   Cas-Gnn\cite{luo2020Cas-Gnn} & - & - &0.872 &0.863 &0.923 &0.047 &0.894 &0.894 &0.937 &0.028 &0.838 &0.843 &0.880 &0.081 &- &- &- &-  \\
   HDFNet\cite{HDFNet-ECCV2020} &91.77 &44.15 &0.879 &0.870 &0.925 &0.046 &0.926 &0.921 &0.970 &0.022 &0.846 &0.868 &0.889 &0.085 &0.886 &0.894 &0.930 &0.048\\
   CoNet\cite{Wei2020CoNet} &20.89 &43.66 &0.851 &0.837 &0.917 &0.056 &0.914 &0.902 &0.948 &0.024 &0.848 &0.852 &0.895 &0.076 &0.860 &0.873 &0.917 &0.058\\
   BBS-Net\cite{fan2020bbsnet} &31.20 &49.77 &0.863 &0.843 &0.914 &0.052 &0.934 &0.928 &0.966 &0.021 &0.835 &0.828 &0.870 &0.092 &0.879 &0.884 &0.922 &0.055\\
   JL-DCF\cite{Fu2020JLDCF} &211.06 &143.52 &0.716 &0.689 &0.818 &0.132&0.928 &0.918 &0.857 &0.021 &0.849 &0.854 &0.887 &0.081 &0.885 &0.894 &0.931 &0.049\\
   SPNet\cite{specificity_rgbd_sod} &175.29 &67.88 &0.871 &0.863 &0.920 &0.044 &\blu{0.945} &\blu{0.946} &\red{0.983} &\red{0.014} &0.855 &0.864 &0.900 &0.071 &0.894 &0.904 &0.933 &0.043\\
   CMINet\cite{cascaded_rgbd_sod} &213.00 &188.12 &0.874 &0.860&0.917 &0.047 &0.940 &0.939 &0.979 &0.016 &0.879 &0.874 &0.913 &{0.061} &0.899 &0.910 &0.939 &{0.040}\\
   DCF\cite{Ji_2021_DCF} &108.60 &53.92 &0.852 &0.829 &0.898 &0.054 &0.916 &0.907 &0.955 &0.023 &0.856 &0.860 &0.903 &0.071 &0.874 &0.886 &0.922 &0.052\\ 
   SPSN \cite{lee2022spsn} &- &- &- &- &- &- &- &- &- &- &- &- &- &- &0.892 &0.900 &0.936 &0.043\\\hline
   \multicolumn{19}{c}{Transformer-based} \\\hline
   VST\cite{Liu_2021_ICCV} &51.33 &53.83 &{0.889} &{0.876} &{0.935} &0.045 &0.943 &0.940 &0.978 &0.017 &{0.882} &\red{0.889} &\red{0.921} &{0.061} &{0.904} &{0.915} &{0.944} &{0.040}\\
   SwinNet-B\cite{liu2021swinnet} &122.2 &199.18 &0.892 &0.879 &0.925 &0.040 &0.914 &0.902 &0.955 &0.025 &0.886 &\red{0.889} &\red{0.921} &\red{0.059} &\blu{0.911} &\red{0.927} &\red{0.950} &\red{0.035}\\
   EBMGSOD\cite{jing_ebm_sod21} &- &- &- &- &- &- &- &- &- &- &0.872 &0.870 &0.906 &0.066 &0.902 &0.918 &0.944 &\blu{0.037}\\
   HRTransNet\cite{tang2022hrtransnet} &18.80 &68.89 &0.839 &0.806 &0.890 &0.059 &0.908 &0.895 &0.938 &0.026 &0.873 &0.871 &0.907 &0.061 &0.859 &0.866 &0.909 &0.059\\
   VST-t\texttt{++} &39.95 &85.40 &\blu{0.896} &0.883 &\blu{0.944} &\blu{0.038} &\blu{0.945} &0.945 &0.981 &\blu{0.015} &0.880 &0.885 &0.916 &0.059 &0.903 &0.917 &\blu{0.946} &0.038\\
   VST-T\texttt{++} &36.95 &100.51 &0.894 &\blu{0.893} &0.939 &0.040 &0.944 &0.943 &0.977 &0.016 &0.883 &\blu{0.887} &\blu{0.919} &\blu{0.060} &0.903 &0.914 &0.944 &0.039\\
   VST-S\texttt{++} &45.41 &143.15 &0.894 &0.887 &0.937 &0.040 &0.944 &0.942&0.979 &0.016 &\red{0.888} &0.879 &0.915 &0.059 &0.904 &0.918 &\blu{0.946} &0.038 \\
   VST-B\texttt{++} &{102.14} &{217.73} &\red{0.915} &\red{0.901} &\red{0.954} &\red{0.030} &\red{0.949} &\red{0.952} &\blu{0.982} &\red{0.014} &\red{0.888} &\blu{0.887} &0.915 &\blu{0.060} &\red{0.912} &\blu{0.925} &\red{0.950} &\red{0.035}\\\hline
  \end{tabular}}
  \label{RGBD_SOTA2}
\end{table*}

\begin{table*}[htbp]
  \centering
  \footnotesize
  \renewcommand{\arraystretch}{1.0}
  \renewcommand{\tabcolsep}{0.7mm}
 \caption{\textbf{Quantitative comparison of our proposed VST\texttt{++} with other 19 SOTA RGB-D SOD methods on five benchmark datasets.}
 Here we employ the average MACs value of our model on all RGB-D datasets because the number of the selected tokens is different for each image. `-' indicates the code or result is not available. `-t', `-T', `-S' and `-B' indicate the T2T-ViT$_t$-14 backbone, the SwinT-1k backbone, the SwinS-1k backbone, and the SwinB-22k backbone with $288\times 228$ image size.
 \red{Red} and \blu{blue} denote the best and the second-best results, respectively.}
  \resizebox{\linewidth}{!}{
  \begin{tabular}{l|cccc|cccc|cccc|cccc|cccc}
  \hline

  \multicolumn{1}{c|}{Summary} & \multicolumn{4}{c|}{NJUD \cite{ju2014njud}} & \multicolumn{4}{c|}{NLPR\cite{peng2014nlpr}} & \multicolumn{4}{c|}{DUTLF-Depth\cite{Piao2019dmra}} & \multicolumn{4}{c|}{ReDWeb-S\cite{liu2021learning}} & \multicolumn{4}{c}{STERE\cite{niu2012stere}} \\
  \hline
   \multicolumn{1}{l|}{Metric}
   & $S_m\uparrow$ & maxF$\uparrow$ & $E_{\xi}^{\text{max}}\uparrow$ & MAE$\downarrow$  & $S_m\uparrow$ & maxF$\uparrow$ & $E_{\xi}^{\text{max}}\uparrow$ & MAE$\downarrow$  & $S_m\uparrow$ & maxF$\uparrow$ & $E_{\xi}^{\text{max}}\uparrow$ & MAE$\downarrow$  & $S_m\uparrow$ & maxF$\uparrow$ & $E_{\xi}^{\text{max}}\uparrow$ & MAE$\downarrow$  & $S_m\uparrow$ & maxF$\uparrow$ & $E_{\xi}^{\text{max}}\uparrow$ & MAE$\downarrow$ \\
   \hline
   \multicolumn{21}{c}{CNN-based} \\\hline
   $S^2$MA\cite{liu2020S2MA} & 0.894 &0.889 &0.930 &0.054 &0.916 &0.902 &0.953 &0.030 &0.904 &0.899 &0.935 &0.043 &0.771 &0.696 &0.781 &0.139 &0.890 &0.882 &0.932 &0.051\\ 
   PGAR\cite{chen2020PGAR} &0.909 &0.907 &0.940 &0.042 &0.917 &0.897 &0.950 &0.027 &0.899 &0.898 &0.933 &0.041 &0.656 &0.632 &0.749 &0.161 &0.894 &0.880 &0.929 &0.045 \\
   DANet\cite{zhao2020DANet} &0.899 &0.898 &0.935 &0.046 &0.920 &0.909 &0.955 &0.027 &0.899 &0.904 &0.939 &0.042 & - & - & - & - &0.901 &0.892 &0.937 &0.044\\
   cmMS\cite{li2020cmMS} &0.900 &0.897 &0.936 &0.044 &0.919 &0.904 &0.955 &0.028 &0.912 &0.913 &0.940 &0.036 &0.699 &0.677 &0.767 &0.143 &0.894 &0.887 &0.936 &0.045\\
   ATST\cite{zhang2020ATSA} &0.885 &0.893 &0.930 &0.047 &0.909 &0.898 &0.951 &0.027 &0.916 &0.928 &0.953 &0.033 &0.679 &0.673 &0.758 &0.155 &0.896 &0.901 &0.942 &0.038\\
   CMW\cite{Li2020CMWNet} &0.870 &0.871 &0.927 &0.061 &0.917 &0.903 &0.951 &0.029 &0.797 &0.779 &0.864 &0.098 &0.634 &0.607 &0.714 &0.195 &0.852 &0.837 &0.907 &0.067\\
   Cas-Gnn\cite{luo2020Cas-Gnn} &0.911 &0.916 &0.948 &0.036 &0.919 &0.906 &0.955 &0.025 &0.920 &0.926 &0.953 &0.030 & - &- &- &- &0.899 &0.901 &0.944 &0.039\\
   HDFNet\cite{HDFNet-ECCV2020} &0.908 &0.911 &0.944 &0.039 &0.923 &0.917 &0.963 &0.023 &0.908 &0.915 &0.945 &0.041 &0.728 &0.717 &0.804 &0.129 &0.900 &0.900 &0.943 &0.042\\
   CoNet\cite{Wei2020CoNet} &0.896 &0.893 &0.937 &0.046 &0.912 &0.893 &0.948 &0.027 &0.923 &0.932 &0.959 &0.029 &0.696 &0.693 &0.782 &0.147 &0.905 &0.901 &0.947 &0.037\\
   BBS-Net\cite{fan2020bbsnet} &0.921 &0.919 &0.949 &0.035 &0.931 &0.918 &0.961 &0.023 &0.882 &0.870 &0.912 &0.058 &0.693 &0.680 &0.763 &0.150 &0.908 &0.903 &0.942 &0.041\\
   JL-DCF\cite{Fu2020JLDCF} &0.877 &0.892 &0.941 &0.066 &0.931 &0.918 &0.965 &0.022 &0.894 &0.891 &0.927 &0.048 &0.581 &0.546 &0.708 &0.213 &0.900 &0.895 &0.942 &0.044\\
   SPNet\cite{specificity_rgbd_sod} &0.925 &0.928 &0.957 &\blu{0.029} &0.927 &0.919 &0.962 &0.021 &0.895 &0.899 &0.933 &0.045 &0.710 &0.715 &0.798 &0.129 &0.907 &0.906 &0.949 &0.037\\
   CMINet\cite{cascaded_rgbd_sod} &\blu{0.929} &\blu{0.934} &0.957 &\blu{0.029} &0.932 &0.922 &0.963 &0.021 &0.912 &0.913 &0.938 &0.038 &0.725 &0.726 &0.800 &0.121 &0.918 &0.916 &0.951 &0.032\\
   DCF\cite{Ji_2021_DCF} &0.904 &0.905 &0.943 &0.039 &0.922&0.910 &0.957 &0.024 &0.925 &0.930 &0.956 &0.030 &0.709 &0.715 &0.790 &0.135 &0.906 &0.904 &0.948 &0.037\\
   SPSN\cite{lee2022spsn} &- &- &- &- &0.923 &0.912 &0.960 &0.023 &- &- &- &- &- &- &- &- &0.907 &0.902 &0.945 &0.036 \\\hline
   \multicolumn{21}{c}{Transformer-based} \\\hline
   VST\cite{Liu_2021_ICCV} &0.922 &0.920 &0.951 &0.035 &0.932 &0.920 &0.962 &0.024 &0.943 &0.948 &\blu{0.969} &0.024 &0.759 &0.763 &0.826 &0.113 &0.913 &0.907 &0.951 &0.038\\
   SwinNet-B\cite{liu2021swinnet} &0.920 &0.924 &0.956 &0.034 &\red{0.941} &\red{0.936} &\red{0.974} &\red{0.018} &0.918 &0.920 &0.949 &0.035 &0.757 &0.756 &0.823 &0.112 &0.919 &\blu{0.918} &\blu{0.956} &0.033\\
   EBMGSOD\cite{jing_ebm_sod21} &- &- &- &- &\blu{0.938} &\blu{0.934} &\blu{0.970} &\red{0.018} &- &- &- &- &- &- &- &- &0.916 &0.916 &\blu{0.956} &\blu{0.032}\\
   HRTransNet\cite{tang2022hrtransnet} &0.908 &0.911 &0.945 &0.037 &0.926 &0.916 &0.964 &0.021 &0.925 &0.930 &0.958 &0.028 &0.714 &0.688 &0.781 &0.137 &0.917 &0.915 &0.955 &\red{0.031}\\
   VST-t\texttt{++} &0.926 &0.927 &0.957 &0.031 &0.934 &0.922 &0.966 &\blu{0.020} &0.943 &0.948 &\blu{0.969} &\blu{0.022} &\blu{0.767} &\blu{0.771} &\blu{0.835} &\blu{0.103} &0.913 &0.911 &0.952 &0.035\\
   VST-T\texttt{++} &0.928 &0.929 &\blu{0.958} &0.031 &0.933 &0.921 &0.964 &0.022 &0.944 &0.948 &\blu{0.969} &0.024 &0.756 &0.757 &0.819 &0.114 &0.916 &0.911 &0.950 &0.037\\
   VST-S\texttt{++} &0.928 &0.928 &0.957 &0.031 &0.935 &0.925 &0.964 &0.021 &\blu{0.945} &\blu{0.950} &\blu{0.969} &0.024 &0.763 &0.766 &0.824 &0.110 &\blu{0.921} &0.916 &0.954 &0.034\\
   VST-B\texttt{++} &\red{0.940} &\red{0.944} &\red{0.968} &\red{0.024} &\blu{0.938} &0.933 &0.968 &\blu{0.020} &\red{0.953} &\red{0.959} &\red{0.974} &\red{0.019} &\red{0.776} &\red{0.778} &\red{0.840} &\red{0.100} &\red{0.926} &\red{0.924} &\red{0.957} &\red{0.031}\\\hline
  \end{tabular}}
  \label{RGBD_SOTA1}
\end{table*}

\subsubsection{Effectiveness of Select-Integrate Attention}
The original VST model and the counterpart with the $\mathcal{L}_{sup}$ loss both use the original self-attention \eqref{self-attention} in the decoder layers, which brings quadratic computational complexity. We further replace the self-attention with our proposed select-integrate attention (SIA) in the last two decoder layers and report the MACs and detection performance.

The results in Table~\ref{ablationTabRGB} and Table~\ref{ablationTabRGBD} show that using SIA achieves comparable or slightly lower detection accuracy on some datasets, mainly due to the coarse information involved in the background area. 
However, despite the existence of this slight performance drop, SIA brings a benefit for approximately 25$\%$ or even more reduction in computation cost. 
This highlights our SIA approach effectively balances accuracy and computational efficiency by discarding less informative background tokens.

\subsubsection{Effectiveness of Depth Position Encoding} 
Considering the peculiarity of depth maps, we design a new depth position encoding (DPE) in addition to the widely used 2D spatial position encoding. The results from Table~\ref{ablationTabRGBD} demonstrate that the use of our proposed DPE yields improved performance on most datasets while bringing negligible MAC increase, compared to relying solely on 2D spatial PEs. Compared to traditional feature-fusion-based RGB-D decoders \cite{liu2020S2MA,li2020cmMS,luo2020Cas-Gnn,Piao2019dmra,HDFNet-ECCV2020}, which usually double the computational costs, our DPE explores a new approach of light-weight depth-aware decoder design.
Please note that the slightly decreased MACs in Table~\ref{ablationTabRGBD} denotes adding DPE leads to more accurate and smaller foreground region segmentation, hence resulting in less selected tokens.

\vspace{3mm}
\subsection{Comparison with State-of-the-Art Methods}
For RGB SOD, we compare our VST\texttt{++} with 10 state-of-the-art CNN-based RGB SOD methods including PiCANet \cite{liu2018picanet}, AFNet \cite{Feng_AFNet}, TSPOANet \cite{Liu_TSPOANet}, EGNet-R \cite{zhao2019EGNet}, ITSD-R \cite{Zhou2020ITSD}, MINet-R \cite{MINet-CVPR2020}, LDF-R \cite{CVPR2020_LDF}, CSF-R2 \cite{{gao2020sod100k}}, GateNet-R \cite{GateNet}, and MENet\cite{wang2023pixels}, as well as three state-of-the-art transformer-based RGB SOD methods: EBMGSOD \cite{jing_ebm_sod21}, ICON \cite{zhuge2021salient}, and our previous VST \cite{Liu_2021_ICCV}. Moreover, we also consider one generic dense predication model, \ie\ Mask2Former-T \cite{cheng2022masked}. Table~\ref{RGB_SOTA} shows the comparison results.

For RGB-D SOD, we adopt 15 state-of-the-art CNN-based RGB-D SOD methods, \ie, $S^2$MA \cite{liu2020S2MA}, PGAR \cite{chen2020PGAR}, DANet \cite{zhao2020DANet}, cmMS \cite{li2020cmMS}, ATSA \cite{zhang2020ATSA}, CMW \cite{Li2020CMWNet}, Cas-Gnn \cite{luo2020Cas-Gnn}, HDFNet \cite{HDFNet-ECCV2020}, CoNet \cite{Wei2020CoNet}, BBS-Net \cite{fan2020bbsnet}, JL-DCF-R \cite{Fu2020JLDCF}, SPNet \cite{specificity_rgbd_sod}, CMINet \cite{cascaded_rgbd_sod}, DCF \cite{Ji_2021_DCF}, and SPSN \cite{lee2022spsn}, as well as four state-of-the-art transformer-based RGB-D SOD models, namely SwinNet \cite{liu2021swinnet}, HRTransNet \cite{tang2022hrtransnet}, EBMGSOD \cite{jing_ebm_sod21}, and our previous VST \cite{Liu_2021_ICCV}, for comparison.
Table~\ref{RGBD_SOTA2} and Table~\ref{RGBD_SOTA1} report the comparison results.
 
Following VST, we leverage the pre-trained T2T-ViT$_t$-14 model \cite{yuan2021tokens} as our backbone to create the VST-t\texttt{++} model.
Moreover, some transformer-based models have been proposed for RGB SOD \cite{jing_ebm_sod21, zhuge2021salient} and RGB-D SOD \cite{liu2021swinnet, jing_ebm_sod21, tang2022hrtransnet} with the Swin Transformer family \cite{liu2021Swin} as the backbone. Following this trend, we explore three Swin Transformer models with different scales, \ie\ SwinT-1k, SwinS-1k, and SwinB-22k \cite{liu2021Swin}, thus obtaining our VST-T\texttt{++}, VST-S\texttt{++}, and VST-B\texttt{++} models.
{As the Swin Transformer family comprises four blocks at 1/4, 1/8, 1/16, and 1/32 scales, respectively, which differs from the T2T-ViT$_t$-14 model, we just upsample the 1/32 features to 1/16 via RT2T, and then concatenate them with the original 1/16 features along the channel dimension. Afterward, we use an MLP to project them back to $d$ and input them to the converter.}
The results show that our VST-t++ model
outperforms our previous VST on most datasets while reducing computation costs in terms of MACs, thus demonstrating the efficacy of our extension.
When switching the backbone from T2T-ViT$_t$-14 \cite{yuan2021tokens} to Swin backbones \cite{liu2021Swin}, we observe gradually improved performance along with the enlarged model scale.

Compared with existing CNN-based methods, especially those widely using the ResNet50 \cite{he2016resnet} backbone, our VST-T\texttt{++} model surpasses them on most RGB and RGB-D SOD datasets with fair comparison, since SwinT demonstrates similar computational complexity to ResNet50 according to \cite{liu2021Swin}.

When compared with Transformer-based methods, to ensure a fair comparison with ICON \cite{zhuge2021salient}, which employs the SwinB-22k backbone
with an input image size of 384, we also attempt changing our input image size from 224 to 384.
However, this would cause significantly higher MACs (247.75G) compared with ICON (52.59G).
Therefore, we reduce our input image size to $288 \times 288$ and the window size in the SwinB backbone to 9. This leads to comparable MACs and Params with ICON and EBMGSOD \cite{jing_ebm_sod21}. However, our VST-B\texttt{++} model outperforms both of them on five out of six RGB datasets.
Table~\ref{RGBD_SOTA2} and Table~\ref{RGBD_SOTA1} show similar results on RGB-D datasets. Compared with the state-of-the-art transformer-based method SwinNet \cite{liu2021swinnet}, our VST-B\texttt{++} model has comparable Params and MACs, while showing better performance on six out of nine datasets.

We additionally evaluate our VST-T++ against the retrained Mask2Former-T \cite{cheng2022masked}, utilizing the same backbone. 
The findings reveal that VST++ outperforms Mask2Former-T, highlighting the strengths of our distinctive design.

\begin{figure*}[t]
  \graphicspath{{Figures/qualitative/}}
  \centering
  \begin{overpic}[width=1\linewidth]{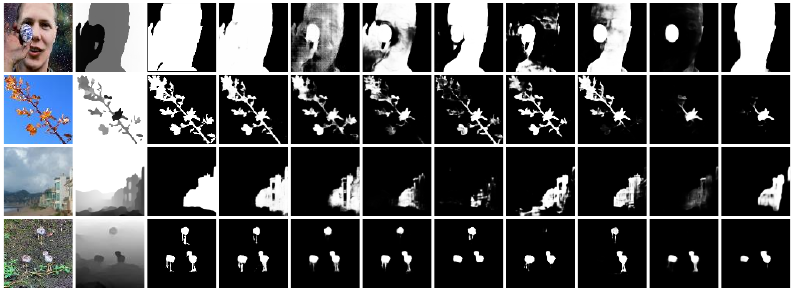}
  \put(2.8,-1.6){\notsotiny Image}
  \put(12.5,-1.6){\notsotiny Depth}
  \put(22.2,-1.6){\notsotiny GT}
  \put(29.3,-1.6){\notsotiny VST-B++}
  \put(39.6,-1.6){\notsotiny VST }
  \put(47.5,-1.6){\notsotiny SwinNet}
  \put(55.2,-1.6){\notsotiny HRTransNet}
  \put(66.0,-1.6){\notsotiny SPNet}
  \put(74.5,-1.6){\notsotiny CMINet}
  \put(83.5,-1.6){\notsotiny BBS-Net}
  \put(93.5,-1.6){\notsotiny DCF}
  \end{overpic}
  \caption{\textbf{Qualitative comparison of our model against state-of-the-art RGB-D SOD methods.} (GT: ground truth.)}
  \label{visualcmpRGBD}
\end{figure*}

\begin{figure*}[t]
  \graphicspath{{Figures/qualitative/}}
  \centering
  \begin{overpic}[width=1\linewidth]{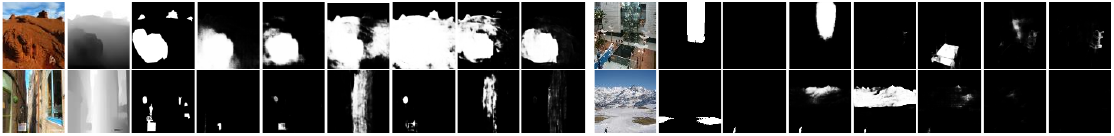}
  \put(1.4,-1.6){\notsotiny Image}
  \put(7.2,-1.6){\notsotiny Depth}
  \put(13.6,-1.6){\notsotiny GT}
  \put(18.0,-1.6){\notsotiny VST-B++}
  \put(25.2,-1.6){\notsotiny VST }
  \put(29.5,-1.6){\notsotiny SwinNet}
  \put(36.0,-1.6){\notsotiny HRTran}
  \put(42.0,-1.6){\notsotiny SPNet}
  \put(47.4,-1.6){\notsotiny CMINet}
  \put(54.2, -1.6){\notsotiny Image}
  \put(60.8,-1.6){\notsotiny GT}
  \put(65.0,-1.6){\notsotiny VST-B++}
  \put(72.0,-1.6){\notsotiny ICON}
  \put(77.0,-1.6){\notsotiny EBMG}
  \put(83.8,-1.6){\notsotiny LDF}
  \put(89.8,-1.6){\notsotiny VST}
 \put(95.5,-1.6){\notsotiny Gate}
  \end{overpic}
  \caption{\textbf{Illustration of failure cases.} Left: RGB-D data. Right: RGB data. (GT: ground truth, EBMG: EBMGSOD, Gate: GateNet, HRTran: HRTransNet.)}
  \label{visualcmpfailure}
\end{figure*}
Figure~\ref{visualcmpRGB} and Figure~\ref{visualcmpRGBD} display visual comparison results among the top-performing models. Our proposed VST\texttt{++} accurately detects salient objects in some challenging situations, \eg, large salient objects, complex background, and multiple salient objects.

\begin{table*}[t]
  \centering
  \footnotesize
  \renewcommand{\arraystretch}{1.0}
  \renewcommand{\tabcolsep}{1mm}
 \caption{\textbf{Quantitative comparison of our proposed VST\texttt{++} with other 14 SOTA RGB, RGB-D, and RGB-T SOD methods on 3 RGB-T benchmark datasets.} `-' indicates the code is not available. \red{Red} and \blu{blue} denote the best and the second-best results, respectively.}
  \begin{tabular}{l|c|c|cccc|cccc|cccc}
  \hline
    \multicolumn{3}{c|}{Summary} &  \multicolumn{4}{c|}{VT821} & \multicolumn{4}{c|}{VT1000} & \multicolumn{4}{c}{VT5000}\\ \hline
    \multicolumn{1}{l|}{Metric} & \multicolumn{1}{c|}{MACs} &\multicolumn{1}{c|}{Params}
    & $S_m$ & maxF & $E_{\xi}^{\text{max}}$ & MAE & $S_m$ & maxF & $E_{\xi}^{\text{max}}$ & MAE & $S_m$ & maxF & $E_{\xi}^{\text{max}}$ & MAE \\\hline
    \multicolumn{15}{c}{RGB $\rightarrow$ RGB-T} \\\hline
    CPD\cite{Wu_CPD} & 17.74 &47.85 &0.784 &0.698 &0.830 &0.083 &0.865 &0.839 &0.913 &0.052 &0.801 &0.735 &0.861 &0.071\\
    BASNet\cite{qin2019basnet} & 97.66 &87.06 &0.814 &0.752 &0.860 &0.067 &0.885 &0.867 &0.923 &0.041 &0.810 &0.741 &0.851 &0.067\\\hline
    \multicolumn{15}{c}{RGB-D $\rightarrow$ RGB-T} \\\hline
    BBSNet\cite{fan2020bbsnet} &28.66 &49.77 &0.869 &0.855 &0.920 &0.048 &0.914 &0.925 &0.964 &0.031 &0.846 &0.825 &0.901 &0.066\\
    TANet\cite{chen2019three} &- &- &0.818 &0.737 &0.856 &0.052 &0.902 &0.884 &0.939 &0.030 &0.847 &0.801 &0.898 &0.046\\
    MMNet\cite{gao2021unified} &- &- &0.875 &0.848 &0.918 &0.039 &0.918 &0.905 &0.953 &0.030&0.864 &0.819 &0.911 &0.046\\\hline
    \multicolumn{15}{c}{CNN-based RGB-T Methods} \\\hline
    SGDL\cite{tu2019rgb} &- &- &0.765 &0.735 &0.839 &0.085 &0.787 &0.770 &0.858 &0.089 &0.750 &0.695 &0.829 &0.088\\
    FCMF\cite{zhang2019rgb} &- &- &0.760 &0.667 &0.810 &0.081 &0.873 &0.851 &0.921 &0.037 &0.814 &0.758 &0.866 &0.055\\
    ADF\cite{tu2022rgbt} &- &- &0.808 &0.749 &0.841 &0.077 &0.909 &0.908 &0.950 &0.034 &0.863 &0.837 &0.911 &0.048\\
    ECFFNet\cite{zhou2021ecffnet} &- &- &0.877 &0.835 &0.911 &0.034 &0.924 &0.919 &0.959 &0.021 &0.876 &0.850 &0.922 &0.037\\
    CGFNet\cite{wang2021cgfnet} &382.63 &69.92 &0.881 &0.866 &0.920 &0.038 &0.923 &0.923 &0.959 &0.023 &0.883 &0.852 &0.926 &0.039\\
    CSRNet\cite{huo2021efficient} &5.76 &1.01 &0.885 &0.855 &0.920 &0.037 &0.919 &0.901 &0.952 &0.027 &0.868 &0.821 &0.912 &0.045\\
    MGAI\cite{song2022multiple} &78.37 &87.09 &0.891 &0.870 &0.933 &0.030 &0.929 &0.921 &0.965 &0.024 &0.884 &0.846 &0.930 &0.037\\
    MIDD\cite{tu2021multi} &217.13 &52.43 &0.871 &0.847 &0.916 &0.044 &0.916 &0.904 &0.956 &0.030 &0.868 &0.834 &0.919 &0.045\\\hline
    \multicolumn{15}{c}{Transformer-based RGB-T Methods} \\\hline
    SwinNet-B\cite{liu2021swinnet} & 122.2 &199.18 &\red{0.904} &\red{0.877} &\red{0.937} &\red{0.029} &\blu{0.938} &\blu{0.933} &\red{0.974} &\red{0.020} &\blu{0.912} &\red{0.885} &\blu{0.944} &\red{0.028}\\
    VST-t\texttt{++} &{39.01} &{85.40} &0.882 &0.843 &0.908 &0.036 &0.936 &0.926 &0.970 &\blu{0.021} &0.891 &0.850 &0.931 &0.037\\
    VST-T\texttt{++} &{35.96} &{100.51} &0.894 &0.861 &0.923 &0.034 &0.941 &0.931 &\blu{0.972} &\red{0.020} &0.895 &0.854 &0.933 &0.037\\
    VST-S\texttt{++} &{44.50} &{143.15} &0.897 &0.868 &0.925 &0.033 &0.940 &0.931 &0.971 &\red{0.020} &0.901 &0.861 &0.936 &\blu{0.034}\\
    VST-B\texttt{++} &102.14 &{217.73} &\red{0.904} &\blu{0.874} &\blu{0.931} &\blu{0.030} &\red{0.943} &\red{0.934} &\red{0.974} &\red{0.020} &\red{0.915} &\blu{0.883} &\red{0.951} &\red{0.028}\\\hline  
  \end{tabular}
  \label{RGBT_SOTA}
\end{table*}

\begin{figure*}[htbp]
  \graphicspath{{Figures/qualitative/}}
  \centering
  \begin{overpic}[width=1\linewidth]{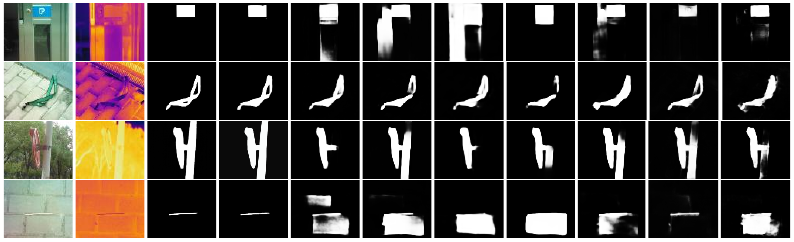}
  \put(2.8,-1.6){\notsotiny Image}
  \put(12.0,-1.6){\notsotiny Thermal}
  \put(22,-1.6){\notsotiny GT}
  \put(29.5,-1.6){\notsotiny VST-B++}
  \put(37.6,-1.6){\notsotiny SwinNet-B}
  \put(48.4,-1.6){\notsotiny MIDD}
  \put(57.4,-1.6){\notsotiny MGAI}
  \put(65.7,-1.6){\notsotiny CSRNet}
  \put(75.2,-1.6){\notsotiny CGFNet}
  \put(84.3,-1.6){\notsotiny ADF}
  \put(92.7,-1.6){\notsotiny MMNet}
  \end{overpic}
  \caption{\textbf{Qualitative comparison of our model against state-of-the-art RGB-T SOD methods.} (GT: ground truth.)}
  \label{visualcmpRGBT}
\end{figure*}

\subsection{Failure Case Analysis}
Although our VST\texttt{++} method outperforms other RGB SOD and RGB-D SOD algorithms and rarely generates completely incorrect prediction results, there are still some failure cases, as depicted in Figure~\ref{visualcmpfailure}. The left part (RGB-D) and right part (RGB) demonstrate cases where the foreground and background are cluttered or similar, making it challenging to define the salient object. In such scenarios, our method consistently highlights the most salient regions. For instance, in the second row of RGB-D examples, which portrays an alley with a messy foreground and background, our method is uncertain whether the board, the lamp, the garnish, and the child all constitute the salient object. Thus, it highlights the most salient region, \ie, the brightest part of the board. Similarly, in the second row of RGB examples, where the foreground and background appear high similarity, our method highlights the most salient people, while the ground truth indicates bare ground. Likewise, other state-of-the-art methods also encounter difficulties with these samples.

\subsection{Application to RGB-T SOD}
To further demonstrate the generalization ability of our proposed VST\texttt{++} model, we apply it to another multi-modality-based SOD task, \ie\ RGB-T SOD, which aims to infer saliency based on a pair of RGB and thermal images. Similar to depth maps in RGB-D SOD, thermal images also provide distinct and complementary information for RGB images, \eg\ penetrating smog and fog. However, thermal images have unique characteristics compared to depth images. Depth images excel at differentiating between foreground and background, whereas thermal images solely focus on discerning thermal properties of a scene, which have less correlation with the foreground and background. Thus, 
we do not employ depth position encoding for thermal images and adopt sinusoidal position encoding instead. 

To ensure fairness in our comparison, we follow previous RGB-T SOD methods \cite{tu2022rgbt,zhou2021ecffnet,wang2021cgfnet,huo2021efficient} and adopt the training set of the VT5000 dataset \cite{tu2022rgbt} to train our model. Subsequently, we evaluate our model on the testing set of VT5000 and two other benchmark datasets, \ie\ VT821 \cite{wang2018rgb}, VT1000 \cite{tu2019rgb}.
For these three benchmark datasets, VT821 \cite{wang2018rgb} contains 821 annotated, yet unregistered image pairs in complex scenes such as low light and multiple salient objects. VT1000 \cite{tu2019rgb} includes 1,000 RGB-T image pairs captured with highly aligned RGB and thermal cameras. VT5000 \cite{tu2022rgbt} is an extensive large-scale dataset with high-resolution, high-diversity, and low-deviation samples, containing 5,000 pairs of RGB-T images.

We compare our models with eight CNN-based state-of-the-art RGB-T SOD models, \ie\ SGDL \cite{tu2019rgb}, FCMF \cite{zhang2019rgb},  ADF \cite{tu2022rgbt}, ECFFNet \cite{zhou2021ecffnet}, CGFNet \cite{wang2021cgfnet}, CSRNet \cite{huo2021efficient}, MGAI \cite{song2022multiple}, and MIDD \cite{tu2021multi}, as well as one state-of-the-art transformer-based RGB-T SOD method: SwinNet \cite{liu2021swinnet}. Additionally, we apply two state-of-the-art RGB SOD models (CPD \cite{Wu_CPD} and BASNet \cite{qin2019basnet}) and three state-of-the-art RGB-D SOD models (BBSNet \cite{fan2020bbsnet}, TANet \cite{chen2019three}, and MMNet \cite{gao2021unified}) to the RGB-T SOD task for comparison, denoted as ``RGB $\rightarrow$ RGB-T" and ``RGB-D $\rightarrow$ RGB-T", respectively.
For RGB SOD models, we keep the models unchanged, but combine the RGB and thermal data as input. In the case of RGB-D SOD models, we substitute the depth inputs with thermal inputs. Their results are provided by \cite{zhou2021ecffnet}.
The comparison results are shown in Table~\ref{RGBT_SOTA}.

The results reveal that our model surpasses all other RGB $\rightarrow$ RGB-T and RGB-D $\rightarrow$ RGB-T models, showcasing the superior generalization ability of our model compared with previous RGB and RGB-D models.
Overall, our VST\texttt{++} is superior to most RGB-T models with comparable computation costs, with the exception on the VT821 dataset. 
Since VT821 includes more noisy samples compared with other datasets, we hypothesize that our pure transformer-based model might be more sensitive to noises compared to CNN-based models. It's worth noting that the SwinNet model's decoder is entirely CNN-based, which could account for its results being less impacted compared to our model.
We offer illustrative instances in Figure~\ref{visualcmpRGBT} to visually compare our method with several state-of-the-art solutions. It is evident that our VST++ highlights salient objects more comprehensively and precisely.

\vspace{3mm}
\section{Conclusion}
This work extends our prior Visual Saliency Transformer (VST) model, which is based on the pure transformer to unify RGB and RGB-D SOD through a sequence-to-sequence task perspective. 
In VST, we designed a multi-task transformer decoder, which allows for the joint execution of saliency and boundary detection in a pure transformer architecture. Additionally, we introduced a novel token upsampling method for transformer-based frameworks, which enables our model to obtain a full-resolution saliency map effortlessly. 
Based on our VST model, in this work, we propose a Select-Integrate Attention (SIA) module to reduce computational costs of traditional self-attention. It selects the foreground regions, splits them into fine-grained segments and aggregates background information into a single token which represents the coarse-grained segments. To incorporate depth position information with low cost, we design a novel depth position encoding method tailored for depth maps. We also introduce an effective token-supervised prediction loss to provide direct supervision signal for saliency and boundary tokens, thus improving model performance. 
To evaluate the effectiveness of our proposed VST\texttt{++}, we conduct comprehensive experiments with diverse transformer-based backbones on RGB, RGB-D, and RGB-T benchmark datasets. Experimental results have exhibited strong generalization ability and improvedd model performance and efficiency of our model.

\bibliographystyle{IEEEtran}
\bibliography{VST_extension1}

\begin{thebibliography}{100}
\providecommand{\url}[1]{#1}
\csname url@samestyle\endcsname
\providecommand{\newblock}{\relax}
\providecommand{\bibinfo}[2]{#2}
\providecommand{\BIBentrySTDinterwordspacing}{\spaceskip=0pt\relax}
\providecommand{\BIBentryALTinterwordstretchfactor}{4}
\providecommand{\BIBentryALTinterwordspacing}{\spaceskip=\fontdimen2\font plus
\BIBentryALTinterwordstretchfactor\fontdimen3\font minus \fontdimen4\font\relax}
\providecommand{\BIBforeignlanguage}[2]{{%
\expandafter\ifx\csname l@#1\endcsname\relax
\typeout{** WARNING: IEEEtran.bst: No hyphenation pattern has been}%
\typeout{** loaded for the language `#1'. Using the pattern for}%
\typeout{** the default language instead.}%
\else
\language=\csname l@#1\endcsname
\fi
#2}}
\providecommand{\BIBdecl}{\relax}
\BIBdecl

\bibitem{shimoda2016distinct}
W.~Shimoda and K.~Yanai, ``Distinct class-specific saliency maps for weakly supervised semantic segmentation,'' in \emph{Eur. Conf. Comput. Vis.}, 2016, pp. 218--234.

\bibitem{zhang2022generalized}
D.~Zhang, G.~Guo, W.~Zeng, L.~Li, and J.~Han, ``Generalized weakly supervised object localization,'' \emph{IEEE Trans Neural Netw Learn Syst}, 2022.

\bibitem{zhao2016person}
R.~Zhao, W.~Oyang, and X.~Wang, ``Person re-identification by saliency learning,'' \emph{IEEE Trans. Pattern Anal. Mach. Intell.}, vol.~39, no.~2, pp. 356--370, 2016.

\bibitem{wang2015saliency}
W.~Wang, J.~Shen, and F.~Porikli, ``Saliency-aware geodesic video object segmentation,'' in \emph{IEEE Conf. Comput. Vis. Pattern Recog.}, 2015, pp. 3395--3402.

\bibitem{lu2020zero}
X.~Lu, W.~Wang, J.~Shen, D.~Crandall, and J.~Luo, ``Zero-shot video object segmentation with co-attention siamese networks,'' \emph{IEEE Trans. Pattern Anal. Mach. Intell.}, 2020.

\bibitem{wang2020paying}
W.~Wang, J.~Shen, X.~Lu, S.~C. Hoi, and H.~Ling, ``Paying attention to video object pattern understanding,'' \emph{IEEE Trans. Pattern Anal. Mach. Intell.}, vol.~43, no.~7, pp. 2413--2428, 2020.

\bibitem{lecun1998gradient}
Y.~LeCun, L.~Bottou, Y.~Bengio, and P.~Haffner, ``Gradient-based learning applied to document recognition,'' \emph{Proc. IEEE}, vol.~86, no.~11, pp. 2278--2324, 1998.

\bibitem{noh2015learning}
H.~Noh, S.~Hong, and B.~Han, ``Learning deconvolution network for semantic segmentation,'' in \emph{Int. Conf. Comput. Vis.}, 2015, pp. 1520--1528.

\bibitem{ronneberger2015unet}
O.~Ronneberger, P.~Fischer, and T.~Brox, ``U-net: Convolutional networks for biomedical image segmentation,'' in \emph{Med. Image. Comput. Comput. Assist. Interv.}, 2015, pp. 234--241.

\bibitem{simonyan2014vgg}
K.~Simonyan and A.~Zisserman, ``Very deep convolutional networks for large-scale image recognition,'' in \emph{Int. Conf. Learn. Represent.}, 2015.

\bibitem{he2016resnet}
K.~He, X.~Zhang, S.~Ren, and J.~Sun, ``Deep residual learning for image recognition,'' in \emph{IEEE Conf. Comput. Vis. Pattern Recog.}, 2016, pp. 770--778.

\bibitem{hou2018dss}
Q.~Hou, M.~Cheng, X.~Hu, A.~Borji, Z.~Tu, and P.~Torr, ``Deeply supervised salient object detection with short connections.'' \emph{IEEE Trans. Pattern Anal. Mach. Intell.}, vol.~41, no.~4, pp. 815--828, 2018.

\bibitem{MINet-CVPR2020}
Y.~Pang, X.~Zhao, L.~Zhang, and H.~Lu, ``Multi-scale interactive network for salient object detection,'' in \emph{IEEE Conf. Comput. Vis. Pattern Recog.}, 2020, pp. 9413--9422.

\bibitem{fan2020bbsnet}
D.-P. Fan, Y.~Zhai, A.~Borji, J.~Yang, and L.~Shao, ``Bbs-net: Rgb-d salient object detection with a bifurcated backbone strategy network,'' in \emph{Eur. Conf. Comput. Vis.}, 2020, pp. 275--292.

\bibitem{luo2020Cas-Gnn}
A.~Luo, X.~Li, F.~Yang, Z.~Jiao, H.~Cheng, and S.~Lyu, ``Cascade graph neural networks for rgb-d salient object detection,'' in \emph{Eur. Conf. Comput. Vis.}, 2020, pp. 346--364.

\bibitem{wang2018salient}
W.~Wang, J.~Shen, X.~Dong, and A.~Borji, ``Salient object detection driven by fixation prediction,'' in \emph{IEEE Conf. Comput. Vis. Pattern Recog.}, 2018, pp. 1711--1720.

\bibitem{zhang2019capsal}
L.~Zhang, J.~Zhang, Z.~Lin, H.~Lu, and Y.~He, ``Capsal: Leveraging captioning to boost semantics for salient object detection,'' in \emph{IEEE Conf. Comput. Vis. Pattern Recog.}, 2019, pp. 6024--6033.

\bibitem{zhao2019EGNet}
J.-X. Zhao, J.-J. Liu, D.-P. Fan, Y.~Cao, J.~Yang, and M.-M. Cheng, ``Egnet:edge guidance network for salient object detection,'' in \emph{Int. Conf. Comput. Vis.}, 2019, pp. 8779--8788.

\bibitem{CVPR2020_LDF}
J.~Wei, S.~Wang, Z.~Wu, C.~Su, Q.~Huang, and Q.~Tian, ``Label decoupling framework for salient object detection,'' in \emph{IEEE Conf. Comput. Vis. Pattern Recog.}, 2020, pp. 13\,025--13\,034.

\bibitem{Wei2020CoNet}
W.~Ji, J.~Li, M.~Zhang, Y.~Piao, and H.~Lu, ``Accurate rgb-d salient object detection via collaborative learning,'' in \emph{Eur. Conf. Comput. Vis.}, 2020, pp. 52--69.

\bibitem{liu2018picanet}
N.~Liu, J.~Han, and M.-H. Yang, ``Picanet: Learning pixel-wise contextual attention for saliency detection,'' in \emph{IEEE Conf. Comput. Vis. Pattern Recog.}, 2018, pp. 3089--3098.

\bibitem{zhang2018pagr}
X.~Zhang, T.~Wang, J.~Qi, H.~Lu, and G.~Wang, ``Progressive attention guided recurrent network for salient object detection,'' in \emph{IEEE Conf. Comput. Vis. Pattern Recog.}, 2018, pp. 714--722.

\bibitem{chen2020dpanet}
Z.~Chen, R.~Cong, Q.~Xu, and Q.~Huang, ``Dpanet: Depth potentiality-aware gated attention network for rgb-d salient object detection,'' \emph{IEEE Trans. Image Process.}, 2020.

\bibitem{Li2020CMWNet}
G.~Li, Z.~Liu, L.~Ye, Y.~Wang, and H.~Ling, ``Cross-modal weighting network for rgb-d salient object detection,'' in \emph{Eur. Conf. Comput. Vis.}, 2020, pp. 665--681.

\bibitem{zhang2020ATSA}
M.~Zhang, S.~X. Fei, J.~Liu, S.~Xu, Y.~Piao, and H.~Lu, ``Asymmetric two-stream architecture for accurate rgb-d saliency detection,'' in \emph{Eur. Conf. Comput. Vis.}, 2020, pp. 374--390.

\bibitem{han2017cnns}
J.~Han, H.~Chen, N.~Liu, C.~Yan, and X.~Li, ``Cnns-based rgb-d saliency detection via cross-view transfer and multiview fusion,'' \emph{IEEE Trans. Cybern.}, vol.~48, no.~11, pp. 3171--3183, 2017.

\bibitem{chen2018progressively}
H.~Chen and Y.~Li, ``Progressively complementarity-aware fusion network for rgb-d salient object detection,'' in \emph{IEEE Conf. Comput. Vis. Pattern Recog.}, 2018, pp. 3051--3060.

\bibitem{Fu2020JLDCF}
K.~Fu, D.-P. Fan, G.-P. Ji, and Q.~Zhao, ``Jl-dcf: Joint learning and densely-cooperative fusion framework for rgb-d salient object detection,'' in \emph{IEEE Conf. Comput. Vis. Pattern Recog.}, 2020, pp. 3052--3062.

\bibitem{specificity_rgbd_sod}
T.~Zhou, H.~Fu, G.~Chen, Y.~Zhou, D.-P. Fan, and L.~Shao, ``Specificity-preserving rgb-d saliency detection,'' in \emph{Int. Conf. Comput. Vis.}, 2021.

\bibitem{HDFNet-ECCV2020}
Y.~Pang, L.~Zhang, X.~Zhao, and H.~Lu, ``Hierarchical dynamic filtering network for rgb-d salient object detection,'' in \emph{Eur. Conf. Comput. Vis.}, 2020, pp. 235--252.

\bibitem{piao2020a2dele}
Y.~Piao, Z.~Rong, M.~Zhang, W.~Ren, and H.~Lu, ``A2dele: Adaptive and attentive depth distiller for efficient rgb-d salient object detection,'' in \emph{IEEE Conf. Comput. Vis. Pattern Recog.}, 2020, pp. 9060--9069.

\bibitem{wang2019salient1}
W.~Wang, Q.~Lai, H.~Fu, J.~Shen, and H.~Ling, ``{Salient object detection in the deep learning era: An in-depth survey},'' \emph{IEEE Trans. Pattern Anal. Mach. Intell.}, 2021.

\bibitem{zhou2021rgb}
T.~Zhou, D.-P. Fan, M.-M. Cheng, J.~Shen, and L.~Shao, ``Rgb-d salient object detection: A survey,'' \emph{Comput. Vis. Media}, pp. 1--33, 2021.

\bibitem{goferman2011context}
S.~Goferman, L.~Zelnik-Manor, and A.~Tal, ``Context-aware saliency detection,'' \emph{IEEE Trans. Pattern Anal. Mach. Intell.}, vol.~34, no.~10, pp. 1915--1926, 2011.

\bibitem{zhao2015saliency}
R.~Zhao, W.~Ouyang, H.~Li, and X.~Wang, ``Saliency detection by multi-context deep learning,'' in \emph{IEEE Conf. Comput. Vis. Pattern Recog.}, 2015, pp. 1265--1274.

\bibitem{ren2015exploiting}
J.~Ren, X.~Gong, L.~Yu, W.~Zhou, and M.~Ying~Yang, ``Exploiting global priors for rgb-d saliency detection,'' in \emph{IEEE Conf. Comput. Vis. Pattern Recog. Workshops}, 2015, pp. 25--32.

\bibitem{luo2017non}
Z.~Luo, A.~Mishra, A.~Achkar, J.~Eichel, S.~Li, and P.-M. Jodoin, ``Non-local deep features for salient object detection,'' in \emph{IEEE Conf. Comput. Vis. Pattern Recog.}, 2017, pp. 6609--6617.

\bibitem{zhai2006visual}
Y.~Zhai and M.~Shah, ``Visual attention detection in video sequences using spatiotemporal cues,'' in \emph{ACM SIGMM Rec.}, 2006, pp. 815--824.

\bibitem{borji2012exploiting}
A.~Borji and L.~Itti, ``Exploiting local and global patch rarities for saliency detection,'' in \emph{IEEE Conf. Comput. Vis. Pattern Recog.}, 2012, pp. 478--485.

\bibitem{cheng2014global}
M.-M. Cheng, N.~J. Mitra, X.~Huang, P.~H. Torr, and S.-M. Hu, ``Global contrast based salient region detection,'' \emph{IEEE Trans. Pattern Anal. Mach. Intell.}, vol.~37, no.~3, pp. 569--582, 2014.

\bibitem{liu2016dhsnet}
N.~Liu and J.~Han, ``Dhsnet: Deep hierarchical saliency network for salient object detection,'' in \emph{IEEE Conf. Comput. Vis. Pattern Recog.}, 2016, pp. 678--686.

\bibitem{wang2017stagewise}
T.~Wang, A.~Borji, L.~Zhang, P.~Zhang, and H.~Lu, ``A stagewise refinement model for detecting salient objects in images,'' in \emph{Int. Conf. Comput. Vis.}, 2017, pp. 4019--4028.

\bibitem{liu2020S2MA}
N.~Liu, N.~Zhang, and J.~Han, ``Learning selective self-mutual attention for rgb-d saliency detection,'' in \emph{IEEE Conf. Comput. Vis. Pattern Recog.}, 2020, pp. 13\,756--13\,765.

\bibitem{vaswani2017attention}
A.~Vaswani, N.~Shazeer, N.~Parmar, J.~Uszkoreit, L.~Jones, A.~N. Gomez, {\L}.~Kaiser, and I.~Polosukhin, ``Attention is all you need,'' in \emph{Adv. Neural Inform. Process. Syst.}, 2017, pp. 5998--6008.

\bibitem{Liu_2021_ICCV}
N.~Liu, N.~Zhang, K.~Wan, L.~Shao, and J.~Han, ``Visual saliency transformer,'' in \emph{Int. Conf. Comput. Vis.}, October 2021, pp. 4722--4732.

\bibitem{dosovitskiy2020image}
A.~Dosovitskiy, L.~Beyer, A.~Kolesnikov, D.~Weissenborn, X.~Zhai, T.~Unterthiner, M.~Dehghani, M.~Minderer, G.~Heigold, S.~Gelly \emph{et~al.}, ``An image is worth 16x16 words: Transformers for image recognition at scale,'' in \emph{Int. Conf. Learn. Represent.}, 2020.

\bibitem{yuan2021tokens}
L.~Yuan, Y.~Chen, T.~Wang, W.~Yu, Y.~Shi, F.~E. Tay, J.~Feng, and S.~Yan, ``Tokens-to-token vit: Training vision transformers from scratch on imagenet,'' in \emph{Int. Conf. Comput. Vis.}, 2021.

\bibitem{Zhou2020ITSD}
H.~Zhou, X.~Xie, J.-H. Lai, Z.~Chen, and L.~Yang, ``Interactive two-stream decoder for accurate and fast saliency detection,'' in \emph{IEEE Conf. Comput. Vis. Pattern Recog.}, 2020, pp. 9141--9150.

\bibitem{zhang2020select}
M.~Zhang, W.~Ren, Y.~Piao, Z.~Rong, and H.~Lu, ``Select, supplement and focus for rgb-d saliency detection,'' in \emph{IEEE Conf. Comput. Vis. Pattern Recog.}, 2020, pp. 3472--3481.

\bibitem{liu2021Swin}
Z.~Liu, Y.~Lin, Y.~Cao, H.~Hu, Y.~Wei, Z.~Zhang, S.~Lin, and B.~Guo, ``Swin transformer: Hierarchical vision transformer using shifted windows,'' in \emph{Int. Conf. Comput. Vis.}, 2021.

\bibitem{vaswani2021scaling}
A.~Vaswani, P.~Ramachandran, A.~Srinivas, N.~Parmar, B.~Hechtman, and J.~Shlens, ``Scaling local self-attention for parameter efficient visual backbones,'' in \emph{IEEE Conf. Comput. Vis. Pattern Recog.}, 2021, pp. 12\,894--12\,904.

\bibitem{zhang2021multi}
P.~Zhang, X.~Dai, J.~Yang, B.~Xiao, L.~Yuan, L.~Zhang, and J.~Gao, ``Multi-scale vision longformer: A new vision transformer for high-resolution image encoding,'' in \emph{Int. Conf. Comput. Vis.}, 2021, pp. 2998--3008.

\bibitem{fang2023reliable}
C.~Fang, Q.~Wang, L.~Cheng, Z.~Gao, C.~Pan, Z.~Cao, Z.~Zheng, and D.~Zhang, ``Reliable mutual distillation for medical image segmentation under imperfect annotations,'' \emph{IEEE Trans Med Imaging}, 2023.

\bibitem{GateNet}
X.~Zhao, Y.~Pang, L.~Zhang, H.~Lu, and L.~Zhang, ``Suppress and balance: A simple gated network for salient object detection,'' in \emph{Eur. Conf. Comput. Vis.}, 2020, pp. 35--51.

\bibitem{xie2015hed}
S.~Xie and Z.~Tu, ``Holistically-nested edge detection,'' in \emph{IEEE Conf. Comput. Vis. Pattern Recog.}, 2015, pp. 1395--1403.

\bibitem{chen2017deeplab}
L.-C. Chen, G.~Papandreou, I.~Kokkinos, K.~Murphy, and A.~L. Yuille, ``Deeplab: Semantic image segmentation with deep convolutional nets, atrous convolution, and fully connected crfs,'' \emph{IEEE Trans. Pattern Anal. Mach. Intell.}, vol.~40, no.~4, pp. 834--848, 2017.

\bibitem{Piao2019dmra}
Y.~Piao, W.~Ji, J.~Li, M.~Zhang, and H.~Lu, ``Depth-induced multi-scale recurrent attention network for saliency detection,'' in \emph{Int. Conf. Comput. Vis.}, 2019, pp. 7254--7263.

\bibitem{wang2018rfcn}
L.~Wang, L.~Wang, H.~Lu, P.~Zhang, and X.~Ruan, ``Salient object detection with recurrent fully convolutional networks,'' \emph{IEEE Trans. Pattern Anal. Mach. Intell.}, vol.~41, no.~7, pp. 1734--1746, 2018.

\bibitem{deng2018r3net}
Z.~Deng, X.~Hu, L.~Zhu, X.~Xu, J.~Qin, G.~Han, and P.-A. Heng, ``R3net: Recurrent residual refinement network for saliency detection,'' in \emph{Int. Joint Conf. Artif. Intell.}, 2018, pp. 684--690.

\bibitem{liu2019salient}
Z.~Liu, S.~Shi, Q.~Duan, W.~Zhang, and P.~Zhao, ``Salient object detection for rgb-d image by single stream recurrent convolution neural network,'' \emph{Neurocomputing}, vol. 363, pp. 46--57, 2019.

\bibitem{chen2020PGAR}
S.~Chen and Y.~Fu, ``Progressively guided alternate refinement network for rgb-d salient object detection,'' in \emph{Eur. Conf. Comput. Vis.}, 2020, pp. 520--538.

\bibitem{qin2019basnet}
X.~Qin, Z.~Zhang, C.~Huang, C.~Gao, M.~Dehghan, and M.~Jagersand, ``Basnet: Boundary-aware salient object detection,'' in \emph{IEEE Conf. Comput. Vis. Pattern Recog.}, 2019, pp. 7479--7489.

\bibitem{chen2019three}
H.~Chen and Y.~Li, ``Three-stream attention-aware network for rgb-d salient object detection,'' \emph{IEEE Trans. Image Process.}, vol.~28, no.~6, pp. 2825--2835, 2019.

\bibitem{zhao2019contrast}
J.-X. Zhao, Y.~Cao, D.-P. Fan, M.-M. Cheng, X.-Y. Li, and L.~Zhang, ``Contrast prior and fluid pyramid integration for rgbd salient object detection,'' in \emph{IEEE Conf. Comput. Vis. Pattern Recog.}, 2019, pp. 3927--3936.

\bibitem{li2020icnet}
G.~Li, Z.~Liu, and H.~Ling, ``Icnet: Information conversion network for rgb-d based salient object detection,'' \emph{IEEE Trans. Image Process.}, vol.~29, pp. 4873--4884, 2020.

\bibitem{liu2021learning}
N.~Liu, N.~Zhang, L.~Shao, and J.~Han, ``Learning selective mutual attention and contrast for rgb-d saliency detection,'' \emph{IEEE Trans. Pattern Anal. Mach. Intell.}, vol.~44, no.~12, pp. 9026--9042, 2021.

\bibitem{wang2018rgb}
G.~Wang, C.~Li, Y.~Ma, A.~Zheng, J.~Tang, and B.~Luo, ``Rgb-t saliency detection benchmark: Dataset, baselines, analysis and a novel approach,'' in \emph{Int J Image Graph}.\hskip 1em plus 0.5em minus 0.4em\relax Springer, 2018, pp. 359--369.

\bibitem{tu2019m3s}
Z.~Tu, T.~Xia, C.~Li, Y.~Lu, and J.~Tang, ``M3s-nir: Multi-modal multi-scale noise-insensitive ranking for rgb-t saliency detection,'' in \emph{Military Interdepartmental Purchase Request}.\hskip 1em plus 0.5em minus 0.4em\relax IEEE, 2019, pp. 141--146.

\bibitem{tu2019rgb}
Z.~Tu, T.~Xia, C.~Li, X.~Wang, Y.~Ma, and J.~Tang, ``Rgb-t image saliency detection via collaborative graph learning,'' \emph{IEEE Trans. Multimedia}, vol.~22, no.~1, pp. 160--173, 2019.

\bibitem{zhang2019rgb}
Q.~Zhang, N.~Huang, L.~Yao, D.~Zhang, C.~Shan, and J.~Han, ``Rgb-t salient object detection via fusing multi-level cnn features,'' \emph{IEEE Trans. Image Process.}, vol.~29, pp. 3321--3335, 2019.

\bibitem{tu2021multi}
Z.~Tu, Z.~Li, C.~Li, Y.~Lang, and J.~Tang, ``Multi-interactive dual-decoder for rgb-thermal salient object detection,'' \emph{IEEE Trans. Image Process.}, vol.~30, pp. 5678--5691, 2021.

\bibitem{tu2022rgbt}
Z.~Tu, Y.~Ma, Z.~Li, C.~Li, J.~Xu, and Y.~Liu, ``Rgbt salient object detection: A large-scale dataset and benchmark,'' \emph{IEEE Trans. Multimedia}, 2022.

\bibitem{wang2021cgfnet}
J.~Wang, K.~Song, Y.~Bao, L.~Huang, and Y.~Yan, ``Cgfnet: Cross-guided fusion network for rgb-t salient object detection,'' \emph{IEEE Trans. Circ. Syst. Video Technol.}, vol.~32, no.~5, pp. 2949--2961, 2021.

\bibitem{fu2022light}
K.~Fu, Y.~Jiang, G.-P. Ji, T.~Zhou, Q.~Zhao, and D.-P. Fan, ``Light field salient object detection: A review and benchmark,'' \emph{Comput Vis Media}, vol.~8, no.~4, pp. 509--534, 2022.

\bibitem{chen2023fusion}
G.~Chen, H.~Fu, T.~Zhou, G.~Xiao, K.~Fu, Y.~Xia, and Y.~Zhang, ``Fusion-embedding siamese network for light field salient object detection,'' \emph{IEEE Trans Multimedia}, 2023.

\bibitem{liu2024deep}
Y.~Liu, X.~Dong, D.~Zhang, and S.~Xu, ``Deep unsupervised part-whole relational visual saliency,'' \emph{Neurocomputing}, vol. 563, p. 126916, 2024.

\bibitem{liu2022disentangled}
Y.~Liu, D.~Zhang, N.~Liu, S.~Xu, and J.~Han, ``Disentangled capsule routing for fast part-object relational saliency,'' \emph{IEEE Trans Image Process}, vol.~31, pp. 6719--6732, 2022.

\bibitem{carion2020end}
N.~Carion, F.~Massa, G.~Synnaeve, N.~Usunier, A.~Kirillov, and S.~Zagoruyko, ``End-to-end object detection with transformers,'' in \emph{Eur. Conf. Comput. Vis.}, 2020, pp. 213--229.

\bibitem{zhu2020deformable}
X.~Zhu, W.~Su, L.~Lu, B.~Li, X.~Wang, and J.~Dai, ``Deformable detr: Deformable transformers for end-to-end object detection,'' in \emph{Int. Conf. Learn. Represent.}, 2020.

\bibitem{wang2020maxdeeplab}
H.~Wang, Y.~Zhu, H.~Adam, A.~Yuille, and L.-C. Chen, ``Max-deeplab: End-to-end panoptic segmentation with mask transformers,'' in \emph{IEEE Conf. Comput. Vis. Pattern Recog.}, 2021, pp. 5463--5474.

\bibitem{liu2021end}
R.~Liu, Z.~Yuan, T.~Liu, and Z.~Xiong, ``End-to-end lane shape prediction with transformers,'' in \emph{IEEE Winter Conf. Appl. Comput. Vis.}, 2021, pp. 3694--3702.

\bibitem{wang2021pvt}
W.~Wang, E.~Xie, X.~Li, D.-P. Fan, K.~Song, D.~Liang, T.~Lu, P.~Luo, and L.~Shao, ``Pyramid vision transformer: A versatile backbone for dense prediction without convolutions,'' \emph{arXiv preprint arXiv:2102.12122}, 2021.

\bibitem{touvron2020training}
H.~Touvron, M.~Cord, M.~Douze, F.~Massa, A.~Sablayrolles, and H.~J{\'e}gou, ``Training data-efficient image transformers \& distillation through attention,'' in \emph{Int. Conf. Mach. Learn.}, 2021, pp. 10\,347--10\,357.

\bibitem{deng2009imagenet}
J.~Deng, W.~Dong, R.~Socher, L.-J. Li, K.~Li, and L.~Fei-Fei, ``Imagenet: A large-scale hierarchical image database,'' in \emph{IEEE Conf. Comput. Vis. Pattern Recog.}, 2009, pp. 248--255.

\bibitem{beltagy2020longformer}
I.~Beltagy, M.~E. Peters, and A.~Cohan, ``Longformer: The long-document transformer,'' \emph{arXiv preprint arXiv:2004.05150}, 2020.

\bibitem{yang2021focal}
J.~Yang, C.~Li, P.~Zhang, X.~Dai, B.~Xiao, L.~Yuan, and J.~Gao, ``Focal self-attention for local-global interactions in vision transformers,'' \emph{arXiv preprint arXiv:2107.00641}, 2021.

\bibitem{zhuge2021salient}
M.~Zhuge, D.-P. Fan, N.~Liu, D.~Zhang, D.~Xu, and L.~Shao, ``Salient object detection via integrity learning,'' \emph{IEEE Trans. Pattern Anal. Mach. Intell.}, 2022.

\bibitem{jing_ebm_sod21}
J.~Zhang, J.~Xie, N.~Barnes, and P.~Li, ``Learning generative vision transformer with energy-based latent space for saliency prediction,'' in \emph{Adv. Neural Inform. Process. Syst.}, 2021.

\bibitem{liu2021swinnet}
Z.~Liu, Y.~Tan, Q.~He, and Y.~Xiao, ``Swinnet: Swin transformer drives edge-aware rgb-d and rgb-t salient object detection,'' \emph{IEEE Trans. Circ. Syst. Video Technol.}, vol.~32, no.~7, pp. 4486--4497, 2021.

\bibitem{tang2022hrtransnet}
B.~Tang, Z.~Liu, Y.~Tan, and Q.~He, ``Hrtransnet: Hrformer-driven two-modality salient object detection,'' \emph{IEEE Trans. Circ. Syst. Video Technol.}, vol.~33, no.~2, pp. 728--742, 2022.

\bibitem{yuan2021hrformer}
Y.~Yuan, R.~Fu, L.~Huang, W.~Lin, C.~Zhang, X.~Chen, and J.~Wang, ``Hrformer: High-resolution vision transformer for dense predict,'' \emph{Adv. Neural Inform. Process. Syst.}, vol.~34, pp. 7281--7293, 2021.

\bibitem{zhang2021k}
W.~Zhang, J.~Pang, K.~Chen, and C.~C. Loy, ``K-net: Towards unified image segmentation,'' \emph{Adv. Neural Inf. Process. Syst.}, vol.~34, pp. 10\,326--10\,338, 2021.

\bibitem{dong2021solq}
B.~Dong, F.~Zeng, T.~Wang, X.~Zhang, and Y.~Wei, ``Solq: Segmenting objects by learning queries,'' \emph{Adv. Neural Inform. Process. Syst.}, vol.~34, pp. 21\,898--21\,909, 2021.

\bibitem{yu2022cmt}
Q.~Yu, H.~Wang, D.~Kim, S.~Qiao, M.~Collins, Y.~Zhu, H.~Adam, A.~Yuille, and L.-C. Chen, ``Cmt-deeplab: Clustering mask transformers for panoptic segmentation,'' in \emph{IEEE Conf. Comput. Vis. Pattern Recog.}, 2022, pp. 2560--2570.

\bibitem{li2022panoptic}
Z.~Li, W.~Wang, E.~Xie, Z.~Yu, A.~Anandkumar, J.~M. Alvarez, P.~Luo, and T.~Lu, ``Panoptic segformer: Delving deeper into panoptic segmentation with transformers,'' in \emph{IEEE Conf. Comput. Vis. Pattern Recog.}, 2022, pp. 1280--1289.

\bibitem{cheng2022masked}
B.~Cheng, I.~Misra, A.~G. Schwing, A.~Kirillov, and R.~Girdhar, ``Masked-attention mask transformer for universal image segmentation,'' in \emph{IEEE Conf. Comput. Vis. Pattern Recog.}, 2022, pp. 1290--1299.

\bibitem{cheng2021per}
B.~Cheng, A.~Schwing, and A.~Kirillov, ``Per-pixel classification is not all you need for semantic segmentation,'' \emph{Adv. Neural Inform. Process. Syst.}, vol.~34, pp. 17\,864--17\,875, 2021.

\bibitem{he2023fastinst}
J.~He, P.~Li, Y.~Geng, and X.~Xie, ``Fastinst: A simple query-based model for real-time instance segmentation,'' in \emph{IEEE Conf. Comput. Vis. Pattern Recog.}, 2023, pp. 23\,663--23\,672.

\bibitem{zhang2023mp}
H.~Zhang, F.~Li, H.~Xu, S.~Huang, S.~Liu, L.~M. Ni, and L.~Zhang, ``Mp-former: Mask-piloted transformer for image segmentation,'' in \emph{IEEE Conf. Comput. Vis. Pattern Recog.}, 2023, pp. 18\,074--18\,083.

\bibitem{ba2016layer}
J.~L. Ba, J.~R. Kiros, and G.~E. Hinton, ``Layer normalization,'' \emph{arXiv preprint arXiv:1607.06450}, 2016.

\bibitem{wang2017duts}
L.~Wang, H.~Lu, Y.~Wang, M.~Feng, D.~Wang, B.~Yin, and X.~Ruan, ``Learning to detect salient objects with image-level supervision,'' in \emph{IEEE Conf. Comput. Vis. Pattern Recog.}, 2017, pp. 136--145.

\bibitem{yan2013ECSSD}
Q.~Yan, L.~Xu, J.~Shi, and J.~Jia, ``Hierarchical saliency detection,'' in \emph{IEEE Conf. Comput. Vis. Pattern Recog.}, 2013, pp. 1155--1162.

\bibitem{li2015HKUIS}
G.~Li and Y.~Yu, ``Visual saliency based on multiscale deep features,'' in \emph{IEEE Conf. Comput. Vis. Pattern Recog.}, 2015, pp. 5455--5463.

\bibitem{li2014PASCALS}
Y.~Li, X.~Hou, C.~Koch, J.~M. Rehg, and A.~L. Yuille, ``The secrets of salient object segmentation,'' in \emph{CVPR}, 2014, pp. 280--287.

\bibitem{everingham2010pascal}
M.~Everingham, L.~Van~Gool, C.~K. Williams, J.~Winn, and A.~Zisserman, ``The pascal visual object classes (voc) challenge,'' \emph{Int J Comput Vis}, vol.~88, pp. 303--338, 2010.

\bibitem{yang2013DUTO}
C.~Yang, L.~Zhang, H.~Lu, X.~Ruan, and M.-H. Yang, ``Saliency detection via graph-based manifold ranking,'' in \emph{IEEE Conf. Comput. Vis. Pattern Recog.}, 2013, pp. 3166--3173.

\bibitem{movahedi2010SOD}
V.~Movahedi and J.~H. Elder, ``Design and perceptual validation of performance measures for salient object segmentation,'' in \emph{IEEE Conf. Comput. Vis. Pattern Recog. Workshops}, 2010, pp. 49--56.

\bibitem{niu2012stere}
Y.~Niu, Y.~Geng, X.~Li, and F.~Liu, ``Leveraging stereopsis for saliency analysis,'' in \emph{IEEE Conf. Comput. Vis. Pattern Recog.}, 2012, pp. 454--461.

\bibitem{ju2014njud}
R.~Ju, L.~Ge, W.~Geng, T.~Ren, and G.~Wu, ``Depth saliency based on anisotropic center-surround difference,'' in \emph{IEEE Int. Conf. Image Process.}, 2014, pp. 1115--1119.

\bibitem{peng2014nlpr}
H.~Peng, B.~Li, W.~Xiong, W.~Hu, and R.~Ji, ``Rgbd salient object detection: A benchmark and algorithms,'' in \emph{Eur. Conf. Comput. Vis.}, 2014, pp. 92--109.

\bibitem{fan2020SIP}
D.-P. Fan, Z.~Lin, Z.~Zhang, M.~Zhu, and M.-M. Cheng, ``Rethinking rgb-d salient object detection: Models, data sets, and large-scale benchmarks,'' \emph{IEEE Trans. Neural Netw. Learn. Syst.}, vol.~32, no.~5, pp. 2075--2089, 2020.

\bibitem{li2014lfsd}
N.~Li, J.~Ye, Y.~Ji, H.~Ling, and J.~Yu, ``Saliency detection on light field,'' in \emph{IEEE Conf. Comput. Vis. Pattern Recog.}, 2014, pp. 2806--2813.

\bibitem{cheng2014rgbd135}
Y.~Cheng, H.~Fu, X.~Wei, J.~Xiao, and X.~Cao, ``Depth enhanced saliency detection method,'' in \emph{International Conference on Internet Multimedia Computing and Service}, 2014, pp. 23--27.

\bibitem{zhu2017ssd}
C.~Zhu and G.~Li, ``A three-pathway psychobiological framework of salient object detection using stereoscopic technology,'' in \emph{Int. Conf. Comput. Vis. Workshops}, 2017, pp. 3008--3014.

\bibitem{fan2017structure}
D.-P. Fan, M.-M. Cheng, Y.~Liu, T.~Li, and A.~Borji, ``Structure-measure: A new way to evaluate foreground maps,'' in \emph{Int. Conf. Comput. Vis.}, 2017, pp. 4548--4557.

\bibitem{Fan2018Enhanced}
D.-P. Fan, C.~Gong, Y.~Cao, B.~Ren, M.-M. Cheng, and A.~Borji, ``{Enhanced-alignment Measure for Binary Foreground Map Evaluation},'' in \emph{Int. Joint Conf. Artif. Intell.}, 2018, pp. 698--704.

\bibitem{gao2020sod100k}
S.-H. Gao, Y.-Q. Tan, M.-M. Cheng, C.~Lu, Y.~Chen, and S.~Yan, ``Highly efficient salient object detection with 100k parameters,'' in \emph{Eur. Conf. Comput. Vis.}, 2020, pp. 702--721.

\bibitem{paszke2019pytorch}
A.~Paszke, S.~Gross, F.~Massa, A.~Lerer, J.~Bradbury, G.~Chanan, T.~Killeen, Z.~Lin, N.~Gimelshein, L.~Antiga \emph{et~al.}, ``Pytorch: An imperative style, high-performance deep learning library,'' \emph{Adv. Neural Inform. Process. Syst.}, vol.~32, pp. 8026--8037, 2019.

\bibitem{gao2019res2net}
S.-H. Gao, M.-M. Cheng, K.~Zhao, X.-Y. Zhang, M.-H. Yang, and P.~Torr, ``Res2net: A new multi-scale backbone architecture,'' \emph{IEEE Trans. Pattern Anal. Mach. Intell.}, vol.~43, no.~2, pp. 652--662, 2019.

\bibitem{Feng_AFNet}
M.~Feng, H.~Lu, and E.~Ding, ``Attentive feedback network for boundary-aware salient object detection,'' in \emph{IEEE Conf. Comput. Vis. Pattern Recog.}, 2019, pp. 1623--1632.

\bibitem{Liu_TSPOANet}
Y.~Liu, Q.~Zhang, D.~Zhang, and J.~Han, ``Employing deep part-object relationships for salient object detection,'' in \emph{Int. Conf. Comput. Vis.}, 2019, pp. 1232--1241.

\bibitem{wang2023pixels}
Y.~Wang, R.~Wang, X.~Fan, T.~Wang, and X.~He, ``Pixels, regions, and objects: Multiple enhancement for salient object detection,'' in \emph{IEEE Conf. Comput. Vis. Pattern Recognit.}, 2023, pp. 10\,031--10\,040.

\bibitem{zhao2020DANet}
X.~Zhao, L.~Zhang, Y.~Pang, H.~Lu, and L.~Zhang, ``A single stream network for robust and real-time rgb-d salient object detection,'' in \emph{Eur. Conf. Comput. Vis.}, 2020, pp. 646--662.

\bibitem{li2020cmMS}
C.~Li, R.~Cong, Y.~Piao, Q.~Xu, and C.~C. Loy, ``Rgb-d salient object detection with cross-modality modulation and selection,'' in \emph{Eur. Conf. Comput. Vis.}, 2020, pp. 225--241.

\bibitem{cascaded_rgbd_sod}
J.~Zhang, D.-P. Fan, Y.~Dai, X.~Yu, Y.~Zhong, N.~Barnes, and L.~Shao, ``Rgb-d saliency detection via cascaded mutual information minimization,'' in \emph{Int. Conf. Comput. Vis.}, 2021.

\bibitem{Ji_2021_DCF}
W.~Ji, J.~Li, S.~Yu, M.~Zhang, Y.~Piao, S.~Yao, Q.~Bi, K.~Ma, Y.~Zheng, H.~Lu, and L.~Cheng, ``Calibrated rgb-d salient object detection,'' in \emph{IEEE Conf. Comput. Vis. Pattern Recog.}, June 2021, pp. 9471--9481.

\bibitem{lee2022spsn}
M.~Lee, C.~Park, S.~Cho, and S.~Lee, ``Spsn: Superpixel prototype sampling network for rgb-d salient object detection,'' in \emph{Eur. Conf. Comput. Vis.}\hskip 1em plus 0.5em minus 0.4em\relax Springer, 2022, pp. 630--647.

\bibitem{Wu_CPD}
Z.~Wu, L.~Su, and Q.~Huang, ``Cascaded partial decoder for fast and accurate salient object detection,'' in \emph{IEEE Conf. Comput. Vis. Pattern Recog.}, 2019, pp. 3907--3916.

\bibitem{gao2021unified}
W.~Gao, G.~Liao, S.~Ma, G.~Li, Y.~Liang, and W.~Lin, ``Unified information fusion network for multi-modal rgb-d and rgb-t salient object detection,'' \emph{IEEE Trans. Circ. Syst. Video Technol.}, vol.~32, no.~4, pp. 2091--2106, 2021.

\bibitem{zhou2021ecffnet}
W.~Zhou, Q.~Guo, J.~Lei, L.~Yu, and J.-N. Hwang, ``Ecffnet: Effective and consistent feature fusion network for rgb-t salient object detection,'' \emph{IEEE Trans. Circ. Syst. Video Technol.}, vol.~32, no.~3, pp. 1224--1235, 2021.

\bibitem{huo2021efficient}
F.~Huo, X.~Zhu, L.~Zhang, Q.~Liu, and Y.~Shu, ``Efficient context-guided stacked refinement network for rgb-t salient object detection,'' \emph{IEEE Trans. Circ. Syst. Video Technol.}, vol.~32, no.~5, pp. 3111--3124, 2021.

\bibitem{song2022multiple}
K.~Song, L.~Huang, A.~Gong, and Y.~Yan, ``Multiple graph affinity interactive network and a variable illumination dataset for rgbt image salient object detection,'' \emph{IEEE Trans. Circ. Syst. Video Technol.}, 2022.

\end{thebibliography}

\vspace{-1.5em}
\begin{IEEEbiography}[{\includegraphics[width=1in,height=1.25in,clip,keepaspectratio]{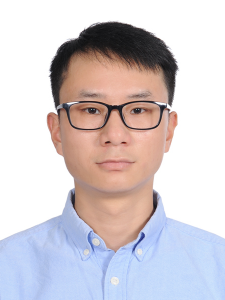}}]
{Nian Liu} is currently a research scientist with Mohamed Bin Zayed University of Artificial Intelligence, Abu Dhabi, UAE. He received the Ph.D. degree and the B.S. degree from the School of Automation at Northwestern Polytechnical University, Xi'an, China, in 2020 and 2012, respectively. His research interests include computer vision and deep learning, especially on saliency detection and few shot learning.
\end{IEEEbiography}

\begin{IEEEbiography}[{\includegraphics[width=1in,height=1.25in,clip,keepaspectratio]{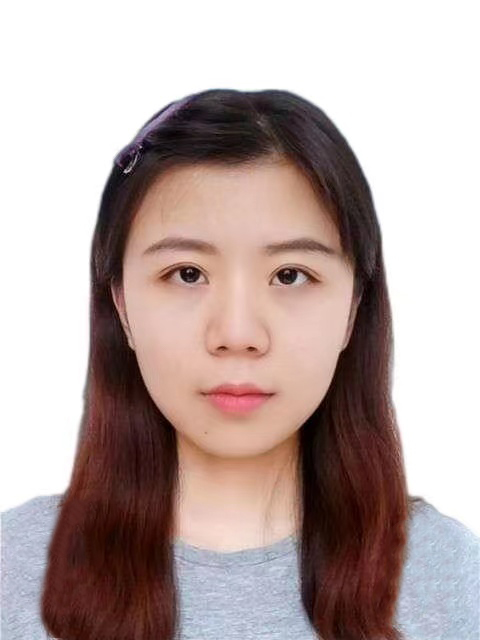}}]
{Ziyang Luo} is a Ph.D. student with the School of Automation, Northwestern Polytechnical University, Xi'an, China. 
She received the B.E. degree in 2022 from Nanjing University of Aeronautics and Astronautics, Nanjing, China. Her research interests include computer vision and deep learning, especially on salient object detection.
\end{IEEEbiography}

\begin{IEEEbiography}[{\includegraphics[width=1in,height=1.25in,clip,keepaspectratio]{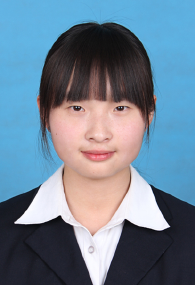}}]
{Ni Zhang} is a Ph.D. student with the School of Automation, Northwestern Polytechnical University, Xi'an, China.
She received the B.E. degree from Northwest A\&F University in 2017. Her research interests include computer vision and deep learning, especially on salient object detection.
\end{IEEEbiography}

\begin{IEEEbiography}[{\includegraphics[width=1in,height=1.25in,clip,keepaspectratio]{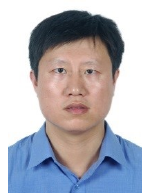}}]
{Junwei Han} is currently a Full Professor with Northwestern Polytechnical University, Xi'an, China. His research interests include computer vision, multimedia processing, and brain imaging analysis. He is an Associate Editor of IEEE Trans. on Pattern Analysis and Machine Intelligence.
\end{IEEEbiography}

\end{document}